\documentclass[sigconf,screen]{acmart}
\usepackage{amsfonts}
\usepackage{amsmath}
\usepackage[utf8x]{inputenc}
\usepackage{multirow}
\usepackage{subcaption}
\usepackage{algorithm}
\usepackage{algorithmicx}
\usepackage{algpseudocode}
\usepackage{bbold}
\usepackage{xspace}
\usepackage{xcolor}
\usepackage{graphicx}
\usepackage{textcomp}
\usepackage[capitalize]{cleveref}
\usepackage[many]{tcolorbox}
\usepackage{enumitem}
\usepackage{microtype}
\usepackage{hyphenat}
\usepackage{hyperref}
\usepackage{booktabs}
\usepackage[]{nth}
\usepackage{esvect}
\usepackage{listings}
\usepackage{url}
\usepackage{natbib}
\usepackage{balance}

\newcommand{\tool}{\textsc{Plato}\xspace}

\makeatletter
\DeclareRobustCommand\onedot{\futurelet\@let@token\@onedot}
\def\@onedot{\ifx\@let@token.\else.\null\fi\xspace}
\def\eg{\emph{e.g}\onedot} \def\Eg{\emph{E.g}\onedot}
\def\ie{\emph{i.e}\onedot} 
 
\def\etc{\emph{etc}\onedot} 
 
\def\etal{\emph{et al}\onedot}
\makeatother

\renewcommand{\paragraph}[1]{\vskip 0.05in \noindent\textbf{#1.}}

\AtBeginDocument{%
  \providecommand\BibTeX{{%
      \normalfont B\kern-0.5em{\scshape i\kern-0.25em b}\kern-0.8em\TeX}}}

\begin{document}

\title{Cross-Lingual Transfer Learning for Statistical Type Inference (update)}

\author{Zhiming Li}
\affiliation{%
  \institution{Nanyang Technological University}
  \country{Singapore}}
\email{zhiming001@e.ntu.edu.sg}

\author{Xiaofei Xie}\authornote{Xiaofei Xie is the corresponding author.}
\affiliation{%
  \institution{Singapore Management University}
  \country{Singapore}}
\email{xfxie@smu.edu.sg}

\author{Haoliang Li}
\affiliation{%
  \institution{City University of Hong Kong}
  \city{Hong Kong}
  \country{China}}
\email{haoliang.li@cityu.edu.hk}

\author{Zhengzi Xu}
\affiliation{%
  \institution{Nanyang Technological University}
  \country{Singapore}}
\email{zhengzi.xu@ntu.edu.sg}

\author{Yi Li}
\affiliation{%
  \institution{Nanyang Technological University}
  \country{Singapore}}
\email{yi_li@ntu.edu.sg}

\author{Yang Liu}
\affiliation{%
  \institution{Nanyang Technological University}
  \country{Singapore}}
\email{yangliu@ntu.edu.sg}

% \author{\IEEEauthorblockN{Zhiming Li\IEEEauthorrefmark{1},
% Xiaofei Xie\IEEEauthorrefmark{2},Haoliang Li\IEEEauthorrefmark{3},
% Zhengzi Xu\IEEEauthorrefmark{2},
% Yi Li\IEEEauthorrefmark{2},
% Yang Liu\IEEEauthorrefmark{2}}
% \IEEEauthorblockA{Nanyang Technological University,\\
% Singapore\\
% \IEEEauthorrefmark{1}zhiming001@e.ntu.edu.sg,
% \IEEEauthorrefmark{2}\{xfxie, zhengzi.xu, yi\_li, yangliu\}@ntu.edu.sg,
% \IEEEauthorrefmark{3}haoliang.li1991@gmail.com}}

\begin{abstract}
% Deep supervised learning-based techniques have been widely applied to the program analysis tasks,
%in fields such as type inference, fault localization, and code summarization.

Hitherto statistical type inference systems rely thoroughly on supervised
learning approaches, which require laborious manual effort to collect and label large amounts of data.
Most Turing-complete imperative languages share similar control- and data-flow structures,
which make it possible to transfer knowledge learned from one language to another.
In this paper, we propose a cross-lingual transfer learning framework, \tool, for statistical type inference, which allows
 us to leverage prior knowledge learned from the labeled dataset of one language and transfer it to the others, \eg, 
Python to JavaScript, Java to JavaScript, etc.
\tool is powered by a novel kernelized attention mechanism to constrain the attention scope of the backbone Transformer model such that the model is forced to base its prediction on commonly shared features among languages. In addition, we propose the syntax enhancement that augments the learning on the feature overlap among language domains.
% Furthermore, \tool can also be used to improve the performance of the conventional supervised learning-based type inference by introducing cross-lingual augmentation, which enables the model to learn more general features across multiple languages.
% \zhiming{Our framework not only allows knowledge transfer among languages, but regularizes model from using spurious bias and thus behaves more robustly.} 
We evaluated \tool under two settings: 1) under the cross-domain scenario that the target language data is not labeled or labeled partially, the results show that \tool outperforms the state-of-the-art domain transfer techniques by a large margin, \eg, it improves the Python to TypeScript baseline by +5.40\%@EM, +5.40\%@weighted-F1, and 2) under the conventional monolingual supervised learning based scenario, \tool improves the Python baseline by +4.40\%@EM, +3.20\%@EM (parametric).
% (1) no labelled target language data, and (2) partial labelled target language data.
% Experimental results show that \tool outperforms the baseline methods by a large margin under both settings.

% \fei{Our contribution is not only transfer learning, we also could use the common knowledge to augment the learning on the same language. Do we need to highlight this point?}
% domain adaptation techniques, \eg, for the Python to JavaScript transfer, \tool improves the best baseline model by
% +14.6\%@EM, +18.6\%@weighted-F1. Besides, \tool consistently improves the baseline under
% the second setting, even for the fully supervised learning baseline, \eg, in JavaScript language, \tool improves the supervised learning based approach by +4.5\%@EM and +3.0\%@weighted-F1.

% Besides, by leveraging data from strongly typed languages, \tool improves the perplexity of the backbone cross-programming-language model and the performance of downstream cross-lingual transfer for type inference.
% Experimental results illustrate that our framework significantly improves the transferability over the baseline method by a large margin.
\end{abstract}
% \keywords{transfer learning, domain adaptation, deep learning, graph kernel, type inference}




\maketitle

\section{Introduction}

\looseness=-1
Deep learning (DL) has achieved tremendous success in many applications such as image
classification and audio recognition.
Recently, DL has also been widely applied in software engineering tasks and obtains superior results
over the traditional rule-based approaches, such as clone
detection~\cite{wei2017supervised,white2016deep}, code
summarization~\cite{zhang2020retrieval,allamanis2016convolutional}, code
translation~\cite{lachaux2020unsupervised}, \etc.

To apply deep learning techniques, large amount of labeled data is required for the training of
high-performance neural networks.
However, it is well-known that manual labeling of data samples for deep learning is extremely
laborious and expensive~\cite{krizhevsky2012imagenet}.
It is more challenging for software engineering tasks, since labeling requires considerable domain
knowledge.
Hence, it would be extremely valuable if we are able to learn models for new languages based on existing labeled
data of another language, avoiding the need to invest additional efforts in labeling.

% Unsupervised learning~\cite{} and transfer learning~\cite{} are the two techniques widely
% studied to address the challenge.
% Unsupervised learning tends to be more challenging, because there is no clear objective
% for the learning.
% \yi{reader knows nothing about unsupervised learning at this point. we can leave out unsupervised learning at this point, only mentioning at related work.}
Transfer learning is becoming increasingly popular, where a model
developed for a domain is reused as the starting point for training a model for another similar domain. The key purpose of transfer learning is to learn more general features on the data to improve the generalization in another domain. For example, in natural language processing, some techniques~\cite{lample2019cross,gururangan2020don} have been proposed to transfer the knowledge between two languages (\eg, English and Nepali). Considering the similarities between different programming language, a natural idea is to adapt the model trained from one language to another language based on transfer learning. Although transfer learning has been extensively studied in the fields of computer vision (CV) and natural language processing (NLP), there is still little research on its applications in program analysis tasks.

However, learning from source code is usually more challenging than in other domains such as images and natural languages.
Comparing with other tasks, it is more challenging to capture program semantics with
deep learning, due to the complex program structures, \eg, sequential execution, looping, branching, \etc.
%Even humans are often difficult to understand the program semantics.
The existing study has shown that DL models learning from programs would easily overfit to some
tokens and it is difficult to learn the real program semantics~\cite{yefet2020adversarial}. It is
unclear whether the existing transfer learning techniques on CV and NLP can still work well on
program domain.

In this paper, we study cross-lingual transfer learning for statistical type inference of optionally-typed programming languages, \ie, adapting the type inference tool trained on programs written in one language to
programs in another language. Type inference~\cite{raychev2015predicting,hellendoorn2018deep,allamanis2020typilus} aims to automatically deduce the type of variables or functions in a dynamic
programming language, which is a fundamental program analysis technique used in bug localization,
program understanding, reverse engineering and
de-obfuscation~\cite{hanenberg2014empirical,gao2017type}.
There have already been some recent attempts on DL-based type inference of optionally-typed
languages~\cite{allamanis2020typilus,hellendoorn2018deep}.
These techniques adopt the mono-lingual supervised learning approach, which works on a given set of
labeled data of the same language, while the trained model is known to have limited
transferability
to other datasets.

Motivated by the fact that the data labeling process for entity types in optionally-typed programming languages is not only labor-intensive but also demands significant expertise knowledge. It is of great potential if we were able to leverage existing labeled dataset from another language to warm start a type inference tool for a new optionally-typed language with scarce data. Notice that most Turing-complete imperative languages share similar control- and data-flow
structures (\eg, variable definitions, \emph{if-else} branches, and loops), which makes the transfer of cross-lingual knowledge possible. To this end, we propose \tool, a cross-lingual transfer learning framework, aiming to train
type inference models with better transferability (\ie, learn more general features).
The key insight of improving transferability is to increase attention on \emph{domain-invariant} features while decreasing attention on \emph{domain-specific} features (\ie, language details).
Specifically, we first perform reaching definition analysis to determine how closely related different tokens are in terms of the type inference.
This information together with the syntax information in abstract syntax tree is then encoded as a novel \emph{kernelized attention} mechanism, which is used as the backbone of our novel \emph{kernelized model}. The idea is to constrain the attention scope of variables in a code sequence during training. Besides, we apply a syntax enhancement strategy which uses srcML~\cite{collard2013srcml} meta-grammar representation to enhance the input representation of the model in order to increase the feature overlap among language domains.
Finally, we adopt a $\kappa\textit{-}$ bagging ensemble strategy that combines kernelized model and unkernelized model for the inference. It is to compensate the negative effect of kernelized model on language-specific corner cases.

To evaluate the effectiveness and usefulness of \tool, we conducted experiments on three different scenarios. 1) The target language dataset is not labeled. We adopt \tool on two popular optionally-typed programming languages: Python and TypeScript, \ie, to use the model trained from the labeled dataset in one language to make predication on
the unlabeled data of another language. We compared \tool with three widely used domain adaptation techniques~\cite{tzeng2014deep,ganin2015unsupervised,gururangan2020don}.
2) The target language dataset is partially labeled. 3) \tool can also be used in the conventional mono-lingual supervised learning based setting.
% We evaluated \tool under two settings: (1) no labeled target language data available, (2) partial labeled target language data available.
% For the first setting, we compared our approach with three widely used domain adaptation techniques~\cite{tzeng2014deep,ganin2015unsupervised,gururangan2020don}. For the second setting, we compared our approach with the several state-of-the-art learning based type inference models, \eg~TypeBert\cite{jesse2021learning}, LambdaNet\cite{wei2020lambdanet}, etc.
The results demonstrate that our method significantly outperforms the baseline methods under all settings, \eg, under the first setting, from Python to TypeScript, \tool improves the best domain adaptation baseline performance by +5.40\% and +5.40\% in terms of EM and weighted-F1. For the second setting, \tool consistently excels the baseline model under all ratios of target domain data. And for the third setting, \tool improves the Python supervised baseline by +4.40\%@EM and +3.20\%@EM (parametric).

In summary, we made the following contributions.
\begin{itemize}[topsep=2pt,itemsep=2pt,partopsep=0ex,parsep=0ex,leftmargin=*]
    \item We propose a cross-lingual transfer learning framework for statistical type inference, which is the first of its kind to the best of our knowledge.
    The framework is powered by the kernelized attention mechanism capturing variable type relations and the syntax enhancement techniques to improve the transferability of the model.
    \item  We demonstrate the feasibility of exploiting the similarity/transferability between different languages in supporting cross-lingual program analysis tasks.
    Our work opens up new opportunities for a wide range of learning-based approaches to be further studied in the future, especially to apply transfer learning in software engineering tasks with multiple languages.

    % Although transfer learning has become a very popular topic in AI tasks (\eg, computer vision and NLP), it is rarely studied in cross-lingual programming tasks. This paper is the first to study transfer learning in type inference tasks. The results and tools could provide insightful information for facilitating the research of software engineering community on more cross-lingual learning tasks. We believe the similarity/transferbaility between program languages is a key feature in software analysis tasks and should be further studied.
    \item We conducted extensive experiments to demonstrate the usefulness and effectiveness of our approach on real-world datasets.
    The results show that \tool significantly outperforms other domain adaptation techniques as well as traditional rule-based models.
    \item We demonstrate that \tool can also outperform the mono-lingual supervised learning based baseline methods by learning more unbiased and general features.
    \item  We have made our tool and data available on our website~\cite{cltl4sti}.
\end{itemize}

\section{Methodology}

\begin{figure}[t]
    \centering
    \includegraphics[width=\columnwidth,trim=12 0 62 0,clip]{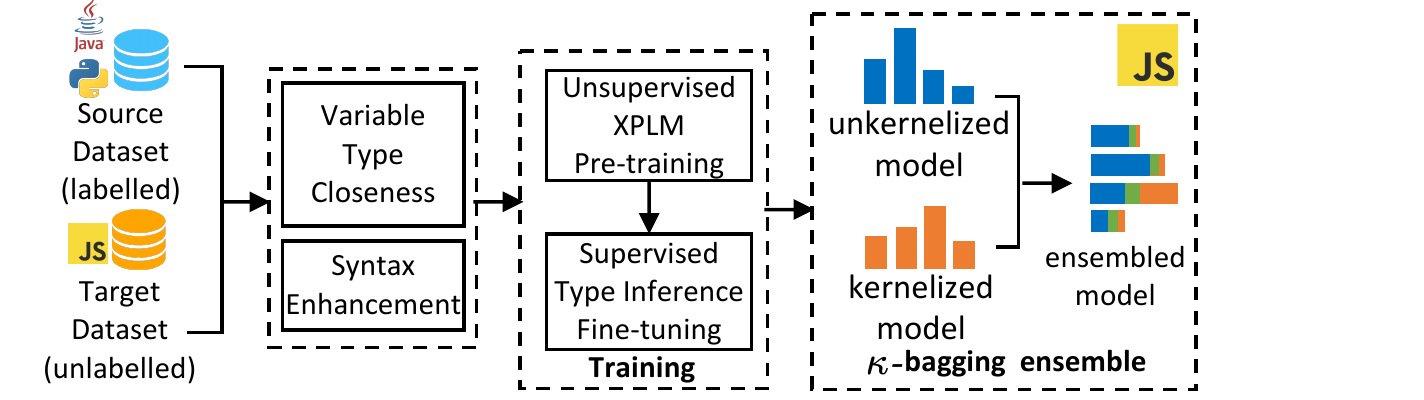}
    \caption{Overview of \tool.}
    \label{fig:overview}
\end{figure}

In this section, we present our framework \tool for cross-lingual transfer learning of statistical type inference in detail.

\subsection{Overview}
%Figure~\ref{fig:overview} shows an overview of our framework, \tool, which aims to train a model
%having high transferability for the type inference task (\eg, from Python to TypeScript).
\Cref{fig:overview} gives an overview of our \tool framework, which consists of four major parts: (1) variable type closeness matrix extraction, (2) syntax enhancement, (3)
training and (4) ensemble-based inference.
The inputs to our system include the source code sequence, its corresponding srcML meta-grammar sequence and variable type closeness matrix. The output is the trained model that can predict the corresponding type annotations for each token in the given code sequence.
%\zhiming{Describe the figure in more detail: what are the inputs, what are in-domain and out-domain
%data, what are the outputs?}

\begin{figure}[t]
\centerline{\includegraphics[width=0.45\textwidth]{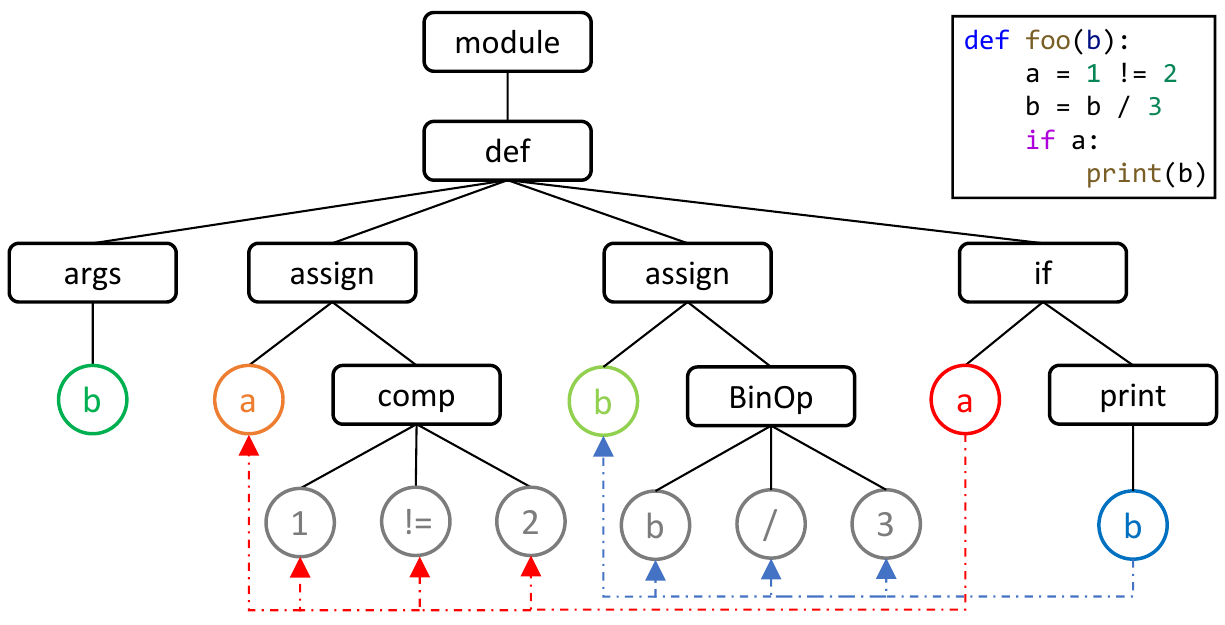}}
\caption{Type-closeness graph of the sample code.}
\label{kernel_img}
\end{figure}

Given an optionally-typed language, our key insights in achieving cross-lingual transfer learning of
type inference are to \emph{exploit the task-relevant features common to the type systems of different
programming languages, and to reduce the impact of the irrelevant features}.
For example, the def-use relationship between variables has a strong connection to their types,
which can be assumed as a common knowledge in many programming languages. %\yi{Up to here.}
For the simple code snippet, ``\texttt{a = 1!=2;~...~print(a)}'', we can infer the type of the
variable ``\texttt{a}'' in the ``\texttt{print}'' statement as Boolean, based on the first statement
``\texttt{a = 1!=2}'' where ``\texttt{a}'' was previously defined.
Such knowledge may seem trivial, but is difficult for deep learning-based models to pick up without
prior knowledge.

To obtain such knowledge, we first perform a reaching definition analysis~\cite{aho2020compilers} on each program for both
the source and target dataset.
%\zhiming{data-flow analysis on AST may not be sound, you mean control-flow graph?}
For each program, based on the result of reaching definition analysis, we define a measurement (\ie, an adjacency matrix) using graph kernel, which we
call \emph{Variable Type Closeness (VTC)} (see \Cref{def:crm}), to quantify the closeness of different tokens in a code sequence in terms of types. %\zhiming{what relationship?}
During training, instead of learning with the traditional attention without constraint, we use the
kernelized attention based on VTC in order to regularize model to focus on the most relevant features for type inference. In this way, the trained model constrains the attention scope of a token only to the tokens related to its type in the sequence and eliminates those that are irrelevant, therefore
it can decrease the negative effect (noise) of the irrelevant features which hinder the transferability.

We further propose a syntax enhancement strategy which is to augment the input representation with srcML meta-grammar~\cite{mikolov2013exploiting,faruqui2014improving}. With srcML meta-grammar representation, features shared between different language domains are augmented such that common semantics can be learnt.
% In essence, the idea is to unify domain-specific keywords unification that normalizes the unique keywords of
% source language
% data and target language data which serve similar syntactic and semantic functionalities into
% unified ones.

One problem is that the kernelized attention model may not be perfect in some predictions. For example, it may overfit to some language-specific features that are mismatched with target languages (see Section~\ref{ensemblesec}).
To mitigate this challenge, we propose an ensemble-based strategy that
combines the kernelized model and un-kernelized model (\ie, attention model learned from code
sequence directly without being constrained by kernel) during the inference.
With such an ensemble strategy, the kernelized and unkernelized models complement each other and produce better results.

\subsection{Variable Type Closeness}
In this part, we introduce the concept of variable type closeness and how it is derived from graph kernel.
% 1) out-domain (Difference with the non-transfer )

% 2) unification (in/out domain)

% \textbf{Variable Relationship,}
% AST+Data-flow Analysis,
% Measurement => Matrix
% equation 3)...

% \subsection{Training}
% Consits: 1) Pretraind,  2) Type superivsed

% a) how to use matrix, above
% b) Loss (see background)

% \subsection{Supervised Type Training?}

% \subsection{Data preprocessing}
% In this section, we introduce the ...
% We apply syntax analysis and data flow analysis in order to obtain the \emph{variable relationship measurement}. Concisely, first, we extract abstract syntax tree for all the samples in the dataset. And upon which, we apply data flow analysis (specifically, reaching definition analysis) and form the joint graph. Finally, based on the joint graph of each sample, we attain the adjacency matrix (\emph{variable relationship measurement}) with a distance metric that models the distance between nodes in the joint graph. The code sequence and its corresponding \emph{variable relationship measurement} are used as the inputs of our model during both the semi-supervised pre-training and supervised type inference training stages.

\paragraph{Kernelized Attention}
For traditional attention mechanism, the embedding of a word depends on its relations with all the
other words in an input sequence, \ie, there is no constraint to its attention scope during gradient-based learning.
For example, consider a code sequence ``\texttt{var~a := true;~ var~b := 0}'', when calculating
embedding for the token ``\texttt{a}'', the traditional attention takes all the other tokens in the
input sequence into consideration.
Yet, when performing type inference for ``\texttt{a}'', the statement ``\texttt{var~b := 0;}'' is
irrelevant and should not be considered when making the prediction.
On the other hand, if the prediction is erroneously based on ``\texttt{b}'' or ``\texttt{0}'', the
model would hardly generalize.
Therefore, we propose a kernelized attention mechanism, which uses a shortest-path graph kernel~\cite{borgwardt2005shortest} to
constrain the attention scope of tokens in code sequence. In this way, given a query token, the model tends to use the set of tokens that are more useful for the type inference, \ie, the closest tokens in the closeness graph for prediction.

To define such a graph kernel, we first introduce a \emph{type-closeness graph} data
structure and define a distance measurement based on the graph.
We then present how to derive the \emph{variable type closeness} adjacency matrix
with an example.

\paragraph{Type-Closeness Graph}
Intuitively, a type closeness graph (TCG) is an annotated AST with extra RDA edges
derived from reaching definition analysis on the Control Flow Graph (CFG).
% where $T$ and $N$ are the terminal and non-terminal nodes in AST, $E_{AST}$ and $E_{RDA}$ are AST
% and RDA edges that connect them,

\begin{definition}[Type-Closeness Graph]
  A \emph{type-closeness graph} is a graph $G=(V, N, E_{AST}, E_{RDA})$, where $V$ and $N$ are
  terminal and non-terminal nodes from the AST, respectively,
  $E_{AST}$ are AST edges, and $E_{RDA}$ contains edges between pairs of terminal nodes $v_i, v_j \in V$ if and only if $v_i$ is within a reaching definition of $v_j$ on the control flow graph.
\end{definition}

\begin{figure}[t]
\centerline{\includegraphics[width=0.25\textwidth]{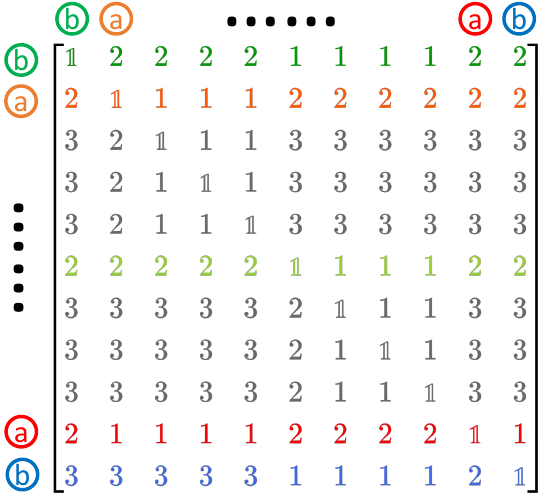}}
\caption{Variable type closeness adjacency matrix $\mathrm{A_{K}^{Q}}$ obtained from
program shown in Figure~\ref{kernel_img}.}\label{adj_mat}
\end{figure}

An example TCG is shown in \Cref{kernel_img},
where the circles represent terminal nodes $V$, the rectangles represent non-terminal
nodes $N$, and the solid (dashed) lines represent $E_{AST}$ ($E_{RDA}$).

\paragraph{Type-Closeness Distance}
\label{tcd}
The \emph{type closeness distance} (TCD) is a distance measure $d(\cdot, \cdot)$
defined over the \emph{type closeness graph}. The smaller the TCD between the target token $v_i$ and the token $v_j$, the more important the token $v_j$ is for the type inference of $v_i$.

\begin{definition}[Type-Closeness Distance]
For a pair of terminal nodes $v_i, v_j \in V$, the type-closeness distance from $v_i$
to $v_j$ is defined as $d(v_{i}, v_{j}) = \min(d_{LCA}(v_{i},v_{j}),
d_{RDA}(v_{i},v_{j}))$,
where $d_{LCA}$ and $d_{RDA}$ are the lowest common ancestor (LCA) distance and the
reaching definition distance, respectively.
\end{definition}

% \fei{
% Intuitively, the type of a variable in a statement (\eg, $x$ in $x = y + `abc'$) usually depends on \textit{other variables in the same statement} (\eg, $y$ and $`abc'$) and \textit{other statements} which have data dependency with the current state (\eg, $y = `zzz'$). To this end, we use the LCA distance $d_{LCA}$ to capture the key tokens in the same statement and the reaching definition distance $d_{RDA}$ to capture the relevant tokens in the data-dependent statements.
% }

% The details of the distance measurement computation are illustrated as follows.
The LCA-distance from $v_i$ to $v_j$ is defined as the length of the shortest path between
$v_i$ and the lowest common ancestor~\cite{aho1976finding} of $v_i$ and $v_j$.
More formally,
\vspace{-5pt}\noindent\par{\small\begin{equation}
d_{LCA}(v_{i}, v_{j}) = d_{AST}(v_{i}, LCA(v_{i},v_{j})),
\label{d_LCA}
\end{equation}}%
where $LCA(\cdot, \cdot)$ denotes the lowest common ancestor of two nodes and $d_{AST}(\cdot, \cdot)$ denotes the distance of the path between two nodes on the AST. For example, as shown in Figure~\ref{kernel_img}, consider node ${\color[RGB]{0,102,204} b}$ and node ${\color[RGB]{255,51,51} a}$, their lowest common ancestor is the non-terminal node \texttt{if}, and it takes two hops from node ${\color[RGB]{0,102,204} b}$ to reach node \texttt{if}, therefore  $d_{LCA}({\color[RGB]{0,102,204} b},{\color[RGB]{255,51,51} a})=2$.
Intuitively, with the $d_{LCA}$, a token is closer to
another token within the
same statement, compared with other tokens from other statements. 
% Thus by regularizing model's learning behavior with $d_{LCA}$, it would be constrained from using tokens from irrelevant statements. 
% \fei{LCA is a bit difficult to understand. For example, LCA is not introduced. SE people does't know LCA. One example is better to explain how to calculate $d_{LCA}$ How to calculate $d_{LCA}$ }

Next, we introduce $d_{RDA}$ that captures the def-use relations between tokens.
Specifically, given a variable node $v_i$, we hard-wire it to the set of nodes $V_{D}=\{v| (v_i, v)\in E_{RDA}\}$ that
comprise its reachable definition statement. For example, the blue dashed lines in Figure~\ref{kernel_img} illustrates the RDA edges of node ${\color[RGB]{0,102,204} b}$.
The RDA-distance $d_{RDA}$ from $v_i$ to all $v_j\in V_{D}$ is defined to be 1, while for others that are unreachable, the distance are set to be $+\infty$.
More formally,%
\vspace{-5pt}\noindent\par{\small\begin{equation}
d_{RDA}(v_{i},v_{j}) = \left\{
             \begin{array}{lr}
             1, & if\ (v_i, v_j) \in E_{RDA} \\
             +\infty,  & otherwise
             \end{array}
\right.
\end{equation}}
Finally, given two nodes $v_i, v_j$ in the TCD space, the type-closeness distance from $v_i$ to $v_j$ is defined as the minimum of their LCA distance and RDA distance: $d(v_{i}, v_{j}) = \min(d_{LCA}(v_{i},v_{j}),
d_{RDA}(v_{i},v_{j}))$.
% The type-closeness distance from $v_i$ to $v_j$ is then the minimum between $d_{LCA}(v_i,v_j)$ and $d_{RDA}(v_i, v_j)$.

% combine RDA之后的distance metric

\paragraph{Variable Type Closeness}
Based on the TCD distance measurement, we derive the \emph{variable type closeness} adjacency matrix, which is used as an input to our model to regularize its learning behavior.

\begin{definition}[Variable Type Closeness]\label{def:crm}
{Given a code sequence $\mathbf{x}$, for each token $t\in \mathbf{x}$,
the \emph{variable type closeness} vector of $t$, denoted as
$\mathbf{A_{\mathbf{x}}^{t}}$, is defined as a distance vector that consists of the distance of $t$
from all the tokens in $\mathbf{x}$ under the TCD defined space, \ie,
$\mathbf{A}_{\mathbf{x}}^{t}=[d(t, t')]_{t'\in \mathbf{x}} \in
\mathbb{R}^{1 \times |\mathbf{x}|}$.
Then by stacking the distance vectors of all tokens within $\mathbf{x}$, forms the \emph{variable type closeness} adjacency matrix of sample $\mathbf{x}$:
$\mathbf{A_{\mathbf{x}}}\in \mathbb{R}^{|\mathbf{x}| \times |\mathbf{x}|}$.}
\end{definition}

\Cref{adj_mat} shows the \emph{variable type closeness} (VTC) adjacency matrix derived
from the TCG graph of the example program $\mathbf{x}$ shown in Figure~\ref{kernel_img}.
The \emph{variable type closeness} distance vector of ${\color[RGB]{0,102,204} b}$: $\mathbf{A}_{\mathbf{x}}^{{\color[RGB]{0,102,204} b}}$ is illustrated in the last row of the matrix. For example, the LCA of ${\color[RGB]{0,102,204} b}$ and ${\color[RGB]{0, 205, 102} b}$ is \texttt{def}, which takes three hops to reach from ${\color[RGB]{0,102,204} b}$ through $E_{AST}$: $d_{LCA}({\color[RGB]{0,102,204} b}, {\color[RGB]{0, 205, 102} b})=3$, and there is no RDA edge that connects them: $d_{RDA}({\color[RGB]{0,102,204} b}, {\color[RGB]{0, 205, 102} b})=+\infty$, thus the first element of its distance vector $\mathbf{A}_{\mathbf{x}}^{{\color[RGB]{0,102,204} b}}[0]=\min(d_{LCA}({\color[RGB]{0,102,204} b}, {\color[RGB]{0, 205, 102} b}),
d_{RDA}({\color[RGB]{0,102,204} b}, {\color[RGB]{0, 205, 102} b}))=3$; and in order to reach ${\color[RGB]{128,128,128} 3}$ in the definition statement, it takes one hop from ${\color[RGB]{0,102,204} b}$ through the $E_{RDA}$: $d_{RDA}({\color[RGB]{0,102,204} b}, {\color[RGB]{128,128,128} 3})=1$ (illustrated as dashed lines), while $d_{LCA}({\color[RGB]{0,102,204} b}, {\color[RGB]{128,128,128} 3})=3$, thus $\mathbf{A}_{\mathbf{x}}^{{\color[RGB]{0,102,204} b}}[8]=\min(d_{LCA}({\color[RGB]{0,102,204} b}, {\color[RGB]{128,128,128} 3}),
d_{RDA}({\color[RGB]{0,102,204} b}, {\color[RGB]{128,128,128} 3}))=1$.

\begin{table}[t]
\caption{\label{tab:anchors}Subset of unified srcML elements.}
\centering
\resizebox{.8\columnwidth}{!}{
\begin{tabular}{l|c|c|c|c}
\toprule
& $Js$     & $Py$     & $Java$    & $srcML$ \\
\midrule
Function definition & function & def & NA     & def      \\
Equality & ===      & ==      & ==      & ==       \\
Non-equality & !==      & !=      & !=      & !=       \\
Logical AND & \&\&     & \&     & \&, \&\& & \&amp      \\
Logical OR & $||$       & $|$       & $|$, $||$   & $|$   \\
Exception & throw    & raise    & throws   & throws    \\
\bottomrule
\end{tabular}}
\end{table}

\subsection{Syntax Enhancement}
For human programmers who manage to master one language, it is relatively easy to switch to
another, because many reserved keywords and operators share the same syntactic and semantics
roles across different language domains.
Deep learning models are hard to exploit this similarity easily, hence limiting the transferability.
To address this, we propose a strategy to augment the syntactic feature overlap shared among
different language domains by incorporating a srcML meta-grammar embedding into the input representation beyond the source code embedding.
At the high level, srcML meta-grammar provides each token in a code sequence with a corresponding markup tag
that represents the abstract syntax role of that token which is unified among languages. \Cref{tab:anchors} shows a subset of the unified tags provided by srcML. Specifically, given a code sequence $\mathbf{x}=(x_1, ..., x_n)$ and its corresponding srcML sequence $\mathbf{s}=(s_1, ..., s_n)$,  we map them to their respective embedding $emb(\mathbf{x})=(emb(x_1), ...emb(x_n))$ and $emb(\mathbf{s})=(emb(s_1), ...emb(s_n))$. Then the augmented input representation $\mathbf{c}$ is the weighted sum of $emb(\mathbf{x})$ and $emb(\mathbf{s})$. Formally:
{\small\begin{equation}
  \mathbf{c}=emb(\mathbf{x})\odot\mathbf{\alpha} + emb(\mathbf{s})\odot\mathbf{\beta}
\label{eq:combo}
\end{equation}}%
% \fei{What is $s$ and how to get $s$? Another language? need to be more clear}
where $\mathbf{\alpha}$ and $\mathbf{\beta}$ are weight vectors for $\mathbf{x}$ and $\mathbf{s}$ respectively, and $\odot$ denotes element-wise multiplication.
% Formally, given the code vector $\mathbf{c}$ and srcML vector $\mathbf{s}$ of an input code sequence, we
% take the weighted sum of them via the element-wise multiplication of $\mathbf{c}$ and $\mathbf{s}$ with
% their corresponding weight vector $\mathbf{\alpha}$ and $\mathbf{\beta}$.
% \fei{What are $K'$ and what is cherry-picked. $K$ has been used before. confusion}
% We initialize the weights for cherry-picked keywords set $\mathbf{K'}$ that serves the same syntactic
% or semantic functionalities to be $\mathbf{\alpha}[\mathbf{k}]=0,
% \mathbf{\beta}[\mathbf{k}]=1~|~\mathbf{k} \in \mathbf{K'}$, which forces the model to leverage the
% srcML token representation instead of the raw code sequence for prediction to achieve higher
% transferability.
In our work, we used srcML to extract the meta grammar representation for Java, and since srcML
does not support Python and TypeScript, we implement an approximate meta grammar mapping for
the two optionally-typed languages on our own.
Our empirical results (see \Cref{sec:rq3-results}) demonstrate that the syntax enhancement
technique is useful in improving the transferability across domains.

%which directly indicates the effectiveness of increasing domain overlap by leveraging the meta grammars.

\subsection{Training}
%We adopt a two-stage training paradigm, including self-supervised cross programming language model pre-training and supervised type inference training.
%During both stages, we use an adjacency matrix for variable relationship measurement purpose to constrain the attention of the model, which we name as kernelized attention.
In this work, we use BERT since it is shown that BERT based model can achieve state-of-the-art performance by leveraging self-supervised pre-training~\cite{jesse2021learning}. Specifically, we use a two-stage training mechanism following \cite{devlin2018bert,wei2020lambdanet}: (1) self-supervised cross programming language model (XPLM)
pre-training, and (2) supervised type inference fine-tuning.
\subsubsection{Unsupervised XPLM Pre-Training.}
\begin{figure}[t]
\centerline{\includegraphics[width=0.33\textwidth,trim=0 7 0 5,clip]{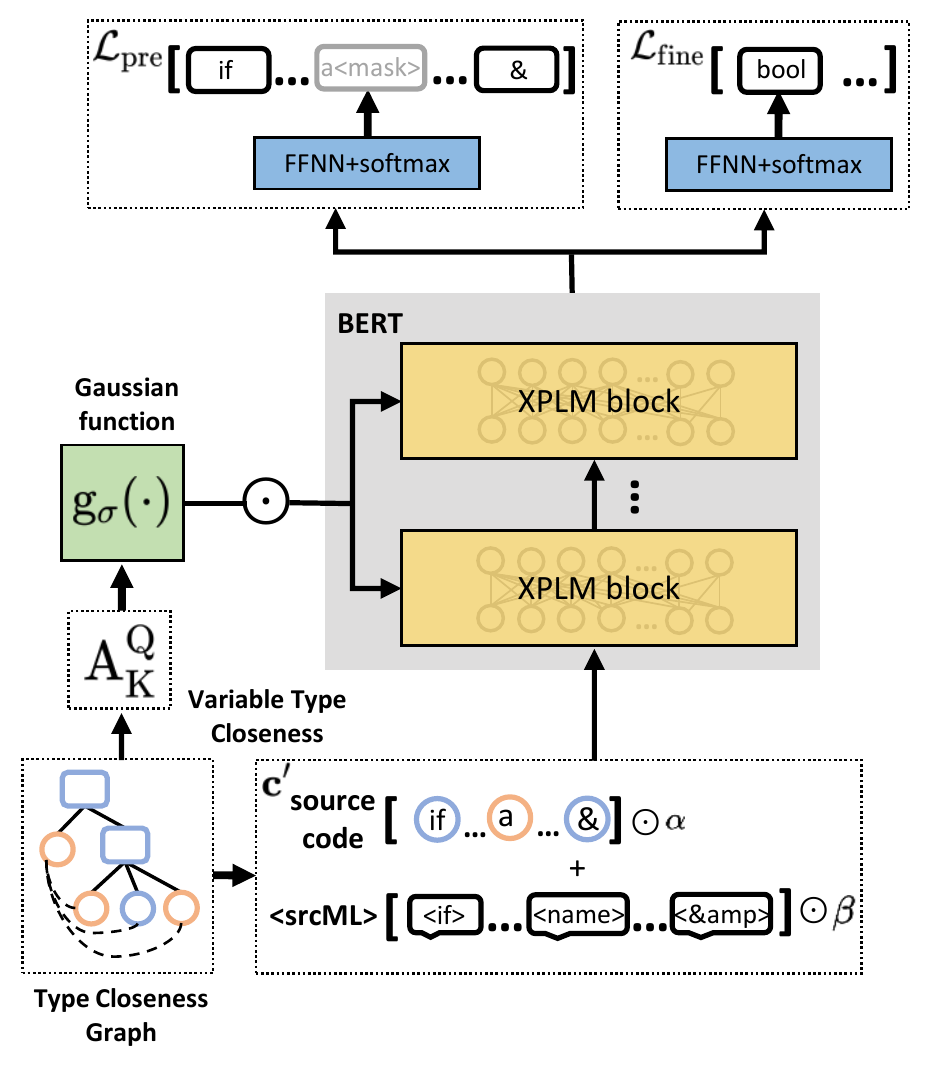}}
\caption{Model architecture.}
\label{model}
\end{figure}

% \zhiming{this subsection is rewritten.}
In this phase, we use data from multiple language sources to pre-train the XPLM model. 
As shown in the model architecture in Figure~\ref{model}, during the self-\\supervised pre-training stage, for each code sequence sample $\mathbf{x}$,
the XPLM backbone model receives two inputs: namely, its corresponding augmented input vector $\mathbf{c}$
and \emph{variable type-closeness} adjacency matrix
$\mathbf{A_{x}}$.
% The latter is used to regularize the attention scope of the model at each layer.
The detailed formulation of the model is given in \Cref{eq:emb}.
\begin{equation}
\mathbf{\kappa{\text -}emb}(\mathbf{x}_i) = \mathbf{g_{\sigma}}(\mathbf{A}_{\mathbf{x} }) \odot
attn(\mathbf{c}_i; \mathbf{c})\cdot \mathbf{c}   
% &= \mathbf{\sum_{k\in \mathbf{K}} \mathbf{g}_{\sigma}(\mathbf{A}_{\mathbf{k}}^{q}) \frac{\langle \mathbf{q,k} \rangle}{\mathbf{Z}}\mathbf{k}} \\
% &= \mathbf{\sum_{k\in \mathbf{K}}} e^{-\frac{||d(\mathbf{q}, \mathbf{k}) - 1||_2^{2}}{2\sigma ^{2}}} \mathbf{\frac{\langle q,k \rangle}{\mathbf{Z}}k}
\label{eq:emb}
\end{equation}

\noindent$\mathbf{c}_i$ is the augmented vector of a token $\mathbf{x}_i$ in the sample code sequence $\mathbf{x}$. We first obtain the attention vector $attn(\mathbf{c}_i;\mathbf{c})\in
\mathbb{R}^{1 \times |\mathbf{x}|}$ of $\mathbf{c}_i$  w.r.t all the vectors in $\mathbf{c}$. Then we constrain the attention by taking element-wise multiplication of $attn(\mathbf{c}_i;\mathbf{c})$ with a regulatory weight vector $\mathbf{g_{\sigma}}(\mathbf{A}_{x})$, where $\mathbf{g_{\sigma}}(\cdot)$ is a radial basis function kernel~\cite{vert2004primer} parameterized by
a learnable or fixed parameter $\mathbf{\sigma}$. Intuitively, the more distant two tokens are in the TCD defined space, the smaller their regulatory weight is. In this way, the model is constrained from using tokens that are irrelevant for embedding. Finally, by taking dot product with $\mathbf{c}$, we obtain the kernelized attention embedding for token $\mathbf{x}_i$. 
We follow the vanilla BERT pre-training paradigm~\cite{devlin2018bert} together with a
regularization loss of $\sigma$ shown as follows:
\vspace{-5pt}\noindent\par{\small\begin{equation}
\label{eq:loss}
\begin{aligned}
\mathcal{L}_{\mathrm{pre} } &= \alpha\mathcal{L}_{\mathrm{MLM} } +\beta\mathcal{L}_{\mathrm{NSP} }+  \gamma\mathcal{L}_{\sigma}\\
\end{aligned}
\end{equation}}
where $\mathcal{L}_{\mathrm{MLM} }$ and $\mathcal{L}_{\mathrm{NSP} }$ denotes the masked language modeling and next sentence prediction loss respectively~\cite{devlin2018bert}. $\alpha, \beta , \gamma$ are regulatory coefficient. The regularization loss of $\mathcal{L}_{\sigma}=\gamma\sigma^{2}$ is used to constrain the attention scope from getting large
during training.
% Specifically, MLM learns token representation by masking out certain proportion of tokens in the sequence and let the model predict for correct words at the masked position. The detailed loss is as follows:
% \begin{equation}
% \begin{aligned}
% \mathcal{L} = \mathcal{L}_{\mathrm{MLM} } = \mathrm{-log(P(x_{[MASK]}|\mathbf{x} \backslash x_{[MASK]}))},
% \end{aligned}
% \end{equation}
% where $\mathbf{x}$ is the entire input code sequence and $x_{[MASK]}$ is the positions that are masked for prediction.

\subsubsection{Supervised Type Inference Fine-Tuning.}\label{sec:finetue}
% \noindent
After obtaining a pre-trained language model from the self-supervised pre-training stage. We fine-tune this model on our downstream type inference task in a supervised manner. In this supervised learning phase, we have the labeled source language samples $S$ and the labeled target language samples $T$. Note that, the number of $T$ is usually small or zero, indicating that we have little or no labeled target language data.

The input of the supervised fine-tuning stage is the same as the pre-training stage, shown in Figure~\ref{model}. 
To allow the model making prediction, we attach a linear layer (FFNN+softmax in \Cref{model}) after the last hidden layer of
the pretrained XPLM to predict the types for each tokens.
We fine-tune all the parameters in the model with a classification loss on the labeled parallel
corpus of code sequence and type annotations.
Specifically, the fine-tune loss function is as follows:
{\small\begin{equation}
\begin{aligned}
\mathcal{L}_{\mathrm{fine} } & = \alpha \mathcal{L}_{\mathrm{fine}}^{S} + \beta \mathcal{L}_{\mathrm{fine}}^{T}\\
&= \alpha\sum_{(\mathbf{x_i, y_i})\in S}\mathbf{-y_{i}} \log [P(\mathbf{\hat{y}_{i}}|\mathbf{x_i})] + \beta\sum_{(\mathbf{x_j, y_j})\in T}\mathbf{-y_{j}} \log [P(\mathbf{\hat{y}_{j}}|\mathbf{x_j})],
\end{aligned}
\end{equation}}
where $\mathcal{L}_{\mathrm{fine}}^{S}$ and $\mathcal{L}_{\mathrm{fine}}^{T}$ are the negative log likelihood loss for samples that are from the source domain and the target domain, respectively. $\mathrm{P(\mathbf{\hat{y}_{i}}|\mathbf{x_i})}$ and $\mathrm{P(\mathbf{\hat{y}_{j}}|\mathbf{x_j})}$ denotes the output probability distribution over the possible type classes for source language sample $\mathbf{x_i}$ and target language sample $\mathbf{x_j}$. %$\mathrm{D_s}$, $\mathrm{D_t}$ denotes the source and target language labeled training data. 
% We embrace a decay training scheme~\cite{ganin2015unsupervised}, $f({\gamma}) = \frac{2}{1+\mathrm{exp(-\gamma)}}-1$ is a regularization term used to control the weight for the loss of the source and target language samples, $\mathrm{\gamma}\in [0,
% 1]$ is the training process.
% Intuitively, when the size of both dataset are imbalanced(\ie, $|S|>>|T|$), we force the model to gradually focus more on the target domain data and less on the source data during the training process in order to mitigate the imbalance problem.
Note that, $S$ usually represent the full source language data that has been labeled, but the size of $T$ could be changed.
Based on the size of $T$, we define different scenarios:
\begin{itemize}[leftmargin=*]
    \item \textit{$|T|=0$ and the model is trained to predict on the target language.} In this setting, all the target training data is not labeled and we conduct the cross-lingual domain adaptation from only the source language.
    \item \textit{$|T|$ is a small number and the model is trained to predict on the target language.} In this setting, a small part of the target language training data (\ie, partially) is labeled and we conduct the cross-lingual transfer learning from source language data as well as the given target language data.
    \item \textit{$|S|$ is a small number and the model is trained to predict on the target language set, with a full size of $|T|$.} This setting corresponds to the common supervised based learning on the target language $T$. The difference is that we also have some labeled training data of other languages (\ie, $S$) beyond the full labeled target language training set. Here, \tool considers $S$ as the augmented data (\ie, the cross-lingual augmentation) and trains a model for the type inference on the language $T$.
\end{itemize}

\begin{algorithm}[t]
  \caption{$\kappa\textit{-}$ bagging BERT}
  \label{alg::ensemleBERT}
  \small
  \begin{algorithmic}[1]
    \Require
    \\
      submodels:\\ $S=\{\mathrm{unkernelized BERT}: \mathrm{BERT}, \mathrm{kernelized BERT}:\mathrm{\kappa \textit{-}BERT}\}$;\\
      dataset: $\mathrm{D=\{x_1, x_2, ...,x_n\}}$;\\
      % confidence threshold: $\theta$;\\
      combination weight: $\lambda$;\\

    \State $D' \leftarrow \emptyset$
    \For {$i \leftarrow 1, 2, ..., |D|$}
    \State obtain logit of each sample from each model: \label{alg:cal10}
    \State $\mathrm{h_{BERT} \leftarrow BERT(x_i)}$ 
    \label{lst:line:l1}
    \State $\mathrm{h_{\kappa \textit{-}BERT} \leftarrow \kappa \textit{-}BERT(x_i)}$ \label{lst:line:l2}
    \State $\mathrm{h_{ensemble} \leftarrow \lambda \cdot h_{\kappa \textit{-}BERT} + (1 - \lambda) \cdot h_{BERT}}$ \label{lst:line:l3}
    \State $\mathrm{D' \cup \{argmax~h_{ensemble}\}}$

    \EndFor

    \Ensure
      \State output ensembled predictions: $\mathrm{D'}$

  \end{algorithmic}
\end{algorithm}
% \zhiming{Specifically, while evaluating under the \emph{no labelelled target language data} setting, $\mathrm{D_s}$ is the full source language labeled training data and $\mathrm{D_t}=\emptyset$; while under \emph{partial labeled target language data} setting, $\mathrm{D_s}$ is a subset of the full source language labeled training data and $\mathrm{D_t}$ would be a subset of the full labeled target language data; finally, when evaluated under the fully supervised learning set, $\mathrm{D_s}$ is a subset of the full source language labeled training data and $\mathrm{D_t}$ would be the full target language labeled training data.} 

%\zhiming{Confused. Is this loss function used to replace (7) or become a part of (7)?}

% Finally, as mentioned above, the kernelized model is used as a specialized submodel for cases with explicit definition statement, so in order to make best use of it, we combine it with a general model trained with traditional attention via an ensemble strategy. The details of the ensemble strategy is illustrated in the next section.

\begin{figure}[t]
%\centerline{\includegraphics[width=0.25\textwidth]{fig/ensemble_se.pdf}}
\begin{tabular}{lcr}
$Js$ Sets &:& \lstinline|var foo = new Set([1, 2, 3])| \\
$Py$ Sets &:& \lstinline|foo = {1, 2, 3}| \\
$Py$ Lists &:& \lstinline|foo = [1, 2, 3]|
\end{tabular}
\caption{An example kernel corner case.}\label{kernel_corner}
\end{figure}

\subsection{Ensemble-Based Inference}\label{ensemblesec}
While the kernelized model is able to use explicit syntactic and semantic relations to improve the performance of type inference, it may fail to cover some corner cases. For example, features within the kernelized attention scope may be language-specific, thus do not generalize to other language domains and lead to negative transfer. As shown in \Cref{kernel_corner}, if we use Python as the source language
and JavaScript as the target language, by using the graph kernel, the XPLM would be
constrained to leverage ``$\texttt{\{$, $\}}$'' and ``$\texttt{[}$, $\texttt{]}$'' as primal features to classify Python
\emph{Sets} and \emph{Lists}.
However, when applied to JavaScript \emph{Sets} sample, the kernelized model would potentially leverage ``$\texttt{[}$,
$\texttt{]}$'' which results in erroneously classifying the variable``$\texttt{foo}$'' into \emph{Lists} instead of
\emph{Sets}.
To this end, we propose an ensemble strategy which combines the kernelized and unkernelized model
during inference stage such that the combined model can make the best of both
worlds.

\Cref{alg::ensemleBERT} shows the detail of the ensemble strategy, which we call $\kappa\textit{-}$ bagging.
Specifically, it is a bagging-based regression ensemble strategy~\cite{breiman1996bagging}.
Given two submodels, the unkernelized model $\mathrm{BERT}$ and the kernelized model $\mathrm{\kappa \textit{-} BERT}$, we first
pass the sample in the test set through both the kernelized and unkernelized models to obtain their corresponding output probability distribution $\mathrm{h_{BERT}}$ and $\mathrm{h_{\kappa \textit{-}BERT}}$ (Line~\ref{lst:line:l1}-\ref{lst:line:l2}). 
% Then we apply an indicator function $\mathbb{1}^{\theta}$ on $\mathrm{h_{BERT}}$
% which returns the original probability distribution vector if its maximum value is larger than the confidence threshold $\theta$ otherwise it returns a zero vector of the same size, such that the ensemble model only uses the prediction of the unkernelized model when it is confident enough.
Then the output ensemble distribution $\mathrm{h_{ensemble}}$ is the weighted sum of $\mathrm{h_{BERT}}$ and $\mathrm{h_{\kappa \textit{-}BERT}}$ using a combination weight $\lambda \in [0, 1]$ (\Cref{lst:line:l3}). 
$\lambda$ is a hyper-parameter that is selected based on the validation set.

\section{Evaluation}
We have implemented \tool based on the PyTorch framework.\footnote{The implementation details and more results can be found on the website~\cite{cltl4sti}.}
To demonstrate the effectiveness and usefulness of \tool in the cross-lingual type inference task, we evaluate
under three settings (see Section~\ref{sec:finetue}):
(1) no labeled target language data available (\textbf{NTL}),
(2) partial labeled target language data available (\textbf{PTL}) and (3) the supervised learning on source language data (\textbf{SL}).
Specifically, we study the following research questions.
\begin{itemize}[topsep=2pt,itemsep=2pt,partopsep=0ex,parsep=0ex,leftmargin=*]
  \item \textbf{RQ1:} How effective is \tool compared with other domain adaptation techniques without any labeled target language data?
  \item \textbf{RQ2:} How do different components of \tool affect the results?
  \item \textbf{RQ3:} How effective is \tool when partial labeled target language data available?
  \item \textbf{RQ4:} How useful is \tool in improving the supervised based baseline methods?
%   \item \textbf{RQ2:} How useful is the kernelized attention in modeling the variable
%   relationship and increase transferability?
%   \item \textbf{RQ3:} How do the proposed augmentation strategies impact the results?
%   \item \textbf{RQ4:} How useful is the ensemble-based inference?
\end{itemize}

\begin{table*}[t]
\centering
\caption{The comparative results with different methods on overlapped types.}
\label{tab:baselinecmp}
\resizebox{.75\textwidth}{!}{
\begin{tabular}{c|cc|cc|cc|cc} 
\toprule
\multirow{2}{*}{Methods}        & \multicolumn{2}{c|}{Python $\rightarrow$ TypeScript} & \multicolumn{2}{c|}{Java $\rightarrow$ TypeScript} & \multicolumn{2}{c|}{TypeScript $\rightarrow$Python} & \multicolumn{2}{c}{Java $\rightarrow$ Python}  \\ 
\cline{2-9}
                                & EM             & weighted-F1                         & EM             & weighted-F1                       & EM             & weighted-F1                        & EM             & weighted-F1                   \\ 
\midrule
TAPT                            & 0.727          & 0.712                               & 0.653          & 0.638                             & 0.550          & 0.565                              & 0.600          & 0.562                         \\
MMD                             & 0.742          & 0.736                               & 0.663          & 0.635                             & 0.590          & 0.572                              & 0.569          & 0.514                         \\
ADV                             & 0.569          & 0.520                               & 0.554          & 0.531                             & 0.531          & 0.492                              & 0.541          & 0.494                         \\
Supervised$_i$                  & 0.713          & 0.707                               & 0.632          & 0.605                             & 0.493          & 0.505                              & 0.572          & 0.516                         \\
\tool                                & \textbf{0.796} & \textbf{0.790}                      & \textbf{0.723} & \textbf{0.684}                    & \textbf{0.633} & \textbf{0.618}                     & \textbf{0.608} & \textbf{0.567}                \\ 
\hline
Improvement ($\Delta$)          & 5.40\%         & 5.40\%                              & 6.00\%         & 4.60\%                            & 4.30\%         & 4.60\%                             & 0.80\%         & 0.50\%                        \\ 
\midrule
\multirow{2}{*}{Supervised$_o$} & \multicolumn{4}{c|}{TypeScript $\rightarrow$TypeScript}                                                   & \multicolumn{4}{c}{Python$\rightarrow$ Python}                                                       \\ 
\cline{2-9}
                                & \multicolumn{2}{c|}{0.886}                           & \multicolumn{2}{c|}{0.885}                         & \multicolumn{2}{c|}{0.688}                          & \multicolumn{2}{c}{0.674}                      \\
\bottomrule
\end{tabular}}
% \label{tab:baselinecmp}
\end{table*}

\subsection{Experimental Setup} \label{sec:setup}

\subsubsection{Dataset Preparation.}
In our experiments, we selected three languages including two optionally-typed languages (\ie, Python and
TypeScript) and one strongly-typed language (\ie, Java).
%We evaluated our method in two settings:
%1) The labels of the source language are available but the labeled target language data is not
%available (\textbf{NTL}) or only partial target language data is available (\textbf{PTL}); 2) The
%labels of the source language and
%(1) the labels of the target language data are not available (\textbf{NTL}), and (2) the labels of partial target language data are available (\textbf{PTL}).
Specifically, for TypeScript, we used a TypeScript dataset~\cite{hellendoorn2018deep} provided by Hellendoorn \etal.
% We generated the corresponding AST of each sample using Esprima~\cite{esprima}.
For Python, we used the dataset provided by Allamanis \etal~\cite{allamanis2020typilus}.
% and extracted corresponding ASTs from their self-defined graphs.
For Java, we used the CodeSearchNet dataset~\cite{husain2019codesearchnet} and extracted the type annotations for variables, function parameters, and return types using srcML~\cite{bui2019sar}.
After the data preprocessing, we collected 20,128 Python
programs with 404,233 variables, 23,267 TypeScript programs with 682,123 variables, and 8,650
Java programs with 213,025 variables, all programs are at function-level. 
% We pre-train the backbone XPLM model using the task-specific datapoints of the above mentioned three languages.

\paragraph{Label Calibration}
The sets of types for different languages may vary.
For example, there are 13,312, 23,047, and 3,810 types in the TypeScript, Python, and Java datasets, respectively.
To facilitate transferability, we need to calibrate the types such that the labels in the training samples (\eg, the source language) and the test samples (\eg, the target language) have the same labels if they have similar functionalities and data structures.
Specifically, we have the following configurations:
\begin{enumerate}[topsep=2pt,itemsep=2pt,partopsep=0ex,parsep=0ex,leftmargin=*]
    \item For \textbf{RQ1} and \textbf{RQ2}, we assume there is not any labeled target language data and aim to
    evaluate the transferability of type system among different languages. Following the similar setting in computer vision
    and natural language processing domains, we relabel the datasets such that the source language
    and the target language have the same co-domain labels set. Specifically, we mainly consider
    the commonly-used types in both source and target languages. Since for TypeScript, there is only one numeric type: \texttt{number},
    thus we consider 4  meta-types \texttt{Boolean}, \texttt{number}, \texttt{string}, \texttt{list}. We use the same set of meta-types for Python.
    % The distribution of meta-types in the three datasets are visualized in \cref{fig:typedistri}.
    \item For \textbf{RQ3}, to simulate the real-life scenario, we assume that there is some target language data that have labels. Here, we consider a more practical setting by using all types (both meta-types and others) in both source language and target language. Suppose $T_s$ and $T_t$ are the set of types in the source language data and the target language data, respectively, the co-domain types we used are $T_s \bigcup T_t$.
    \item For \textbf{RQ4}, we compared \tool with the state-of-the-art supervised based techniques. We keep the prediction type space of our model consistent with the one used in the original baselines.
\end{enumerate}

\subsubsection{Evaluation Measurements.}
\noindent
The data distribution of the type system is imbalanced, \eg,~\texttt{string} takes up a much larger proportion than all the other types in all languages.
Therefore, the widely used \emph{Exact Match} (EM)~\cite{hellendoorn2018deep,allamanis2020typilus}
is suboptimal, because when using EM, a weak classifier that is biased toward predicting types
with the highest occurrences in the training set could still get a spuriously good result.
% \zhiming{I don't understand: it does not take into account the precision and recall rate of different
% classes, which could result in spuriously good result in either precision or recall, but not both.}
Therefore, we use weighted-F1 to account for the precision and recall trade-off as well as the data imbalance.
Formally, weighted-F1 calculates the F1-score for each class and takes their average weighted by support:%
\vspace{-5pt}\noindent\par{\small\begin{equation}
\mathrm{weighted{\text -}F1 =  \sum_{i\in C} \frac{|C_i|}{|C|} \mathrm{F1{\text -}score_{i}}},
\label{weighted}
\end{equation}}%
where $\mathrm{|C_i|}$ is the size of class $\mathrm{C_i}$, and $\mathrm{|C|}$ is the size of the entire dataset.
In our evaluation, we report both the EM values and the weighted-F1 scores.

\subsubsection{Configurations.}
% \fei{@zhiming, add some configuration details, \eg, training parameters and others. }
We used a BERT~\cite{devlin2018bert} encoder with 4 stacked attention layers, 4-headed
attention as the backbone XPLM.
The dimension of all the token embedding is 256. We train the models using Adam
optimizer~\cite{kingma2014adam} with a initial learning rate of $\mathrm{10^{-4}}$.
All models are fine-tuned for 30 epochs and we select the best-performed model on the validation set in terms of EM.
We conducted all experiments on a Ubuntu 16.04 server with 24 cores of 2.2GHz CPU, 251GB RAM
and two GeForce RTX 3090 GPU with a total of 48GB memory.

\subsection{\textbf{RQ1:} Comparison with Baselines When No Labeled Target Data Is Available}

\paragraph{Baselines}
To evaluate the transferability among different languages and effectiveness of our method under the NTL setting, we compared \tool with three popular domain adaptation methods, which are widely used in text and image classification
tasks~\cite{ganin2015unsupervised,tzeng2014deep,gururangan2020don}.
% We re-implemented these methods on our task.

\paragraph{TAPT}
We adopted the Task-Adaptive Pre-Training (TAPT)~\cite{howard2018universal,gururangan2020don},
which leverages the task-relevant data to adapt the pretrained backbone model to
specific downstream domain, as a baseline.
Specifically, TAPT utilizes the unlabeled task-relevant samples from both the source and target domain to further fine-tune the pretrained language model such that it is much more task- and domain-relevant. In this work, we use the whole unlabeled corpus from both the source and target programming languages to adapt the XPLM during pre-training. We use the adapted backbone model for all the baselines and benchmarks in this work.

\paragraph{MMD}
We adopted Maximum Mean Discrepancy (MMD)~\cite{tzeng2014deep,gretton2012kernel} as the second
baseline domain adaptation method.
The key idea of MMD is to minimize the latent feature discrepancy between the source and the target domains such that they become indistinguishable by
the model.
Concisely, MMD attaches a discrepancy loss term on the last hidden layer of the backbone model and maximizes the loss during the type inference training phase.

% As the start of loss-based domain adaptation , Maximum Mean Discrepancy (MMD) loss\cite{tzeng2014deep,gretton2012kernel} is usually used as an additional domain confusion loss and is trained simultaneously with the regular classification loss of the model. Usually, it is directly applied on the latent feature of the models and an optional kernel is implemented to map the latent features to a shared space. In our work, we apply the MMD loss on the last hidden layer of the backbone model and a linear kernel is adopted. \fei{a linear kernel is adopted on what?}

\paragraph{ADV}
For the third baseline, we adopted adversarial domain
adaptation~\cite{ganin2015unsupervised,chen2018adversarial}. ADV transfers knowledge from the
source to the target domain by using reversed gradient drawn from domain classification loss to confuse the
features from the source and target domains.
Specifically, ADV introduces a gradient reversal layer on
top of the last hidden layer of the backbone model. A domain classifier is used to distinguish the domain of samples. The updated gradient from the domain classifier is reversed by the gradient reversal layer before being used to update the model.
% \zhiming{previous sentence too long! hard to parse}

% Beyond the original classification loss, adversarial domain adaptation\cite{ganin2015unsupervised,chen2018adversarial} is achieved by a source-target domain classifier along with a gradient reversal layer, the weights of the backbone network were updated by both the ordinary classification loss and the reversed gradient during the back-propagation. The source and target domains were expected to be indistinguishable after training and the domain-invariant features were therefore extracted.

% \fei{@Haoliang, please check the description on the baselines.}

% \subsection{\textbf{RQ1:} Comparison with baselines when no labeled target data is available}

\paragraph{Setting}
% \subsubsection{Setting}\noindent\label{sec:rq1-setting}
%We compared \tool with three domain adaptation techniques (\ie, TAPT, MMD, and ADV) and one supervised learning technique.
We randomly split the target language dataset into validation-test sets in
15-85 proportions and use the whole source language dataset for training.
%The target training set is used for unsupervised pre-training of XPLM without labels which is included in all domain adaptation techniques.

In addition to the three baselines, we also calculate the results of supervised learning as reference. Specifically, we train a vanilla Transformer classifier with supervised learning on the in-domain source dataset (denoted as Supervised$_i$) and then use the classifier to evaluate the out-domain target dataset without any domain adaptation techniques, which can be regarded as the lower bound of the domain adaptation techniques. On the other hand, we adopt supervised learning to train another Transformer classifier on the out-domain target data and evaluate it on the out-domain dataset (denoted as Supervised$_o$), which can be regarded as the upper bound. For the baseline methods, the regulatory coefficients for MMD and ADV are set to 0.1.

% For the supervised learning baseline, we trained the classifier on either the in-domain or out-domain training set with labels and it is respectively used to evaluate the in-domain and the out-domain test set.

% \subsubsection{Results}\noindent
\paragraph{Results}
\Cref{tab:baselinecmp} shows the detailed results of different methods in
terms of exact match and weighted-F1. Columns show the transfer results from different source
language domains to different target language domains. For example, for the domain adaptation between optionally-typed languages, column ``Python $\rightarrow$
TypeScript'' shows the cross-lingual transfer results from Python to TypeScript.
%while ``TypeScript $\rightarrow$ Python'' shows results from TypeScript to Python.
We also included the results of using the strongly-typed language (\ie, Java) as the source language, which represents the scenario that if we do not have an existing labeled optionally-typed language dataset, we can use the strongly-typed language data as the source because their types can be obtained automatically.

% Note that, on the ``In-domain'' row, we show the results of supervised learning in the same domain,

Overall, the results demonstrate that our method outperforms the three domain
adaptation techniques in terms of all measurements using either optionally-typed language
(TypeScript/Python) or strongly-typed language (Java) as the source language.
Row ``Improvement ($\Delta$)'' shows the improvement of \tool over the best results of the baselines.
Specifically, from Python to TypeScript, the performance of exact match and weighted-F1 is
increased by +5.40\% and +5.40\%, respectively.
From TypeScript to Python, it is increased by +4.30\% and +4.60\%, respectively.
When using Java as the source language: from Java to TypeScript, the results are improved by +6.00\%@EM,
+4.60\%@weighted-F1; from Java to Python, the results are improved by +0.80\%@EM, +0.50\%@weighted-F1.
Interestingly, by using strongly-typed language Java as the source language, we can achieve
comparative performance compared with using optionally-typed language as source language. Furthermore, we compare \tool with the rule-based type inference tool. \Eg~for TypeScript, while TSc+CheckJS\footnote{https://www.typescriptlang.org/tsconfig/checkJs.html} achieves 69.5\%@EM, 81.9\%@weighted-F1, \tool manages to achieve 79.6\%@EM, 79.0\%@weighted-F1. The results show the transferability of the trained model among languages. With \tool, one can achieve comparative or even better performance by using cross-lingual labeled data instead of implementing rule-based tool from scratch that requires significant manual effort and expert knowledge.

Consider the results of supervised learning baseline, not surprisingly, Supervised$_i$ performs poorly on the
out-domain data due to that it does not have any knowledge of the out-domain target data.
Consider Supervised$_o$, we observe that, although our technique has already achieved the best result in the domain adaptation setting, there is still a gap with the supervised learning setting which leaves room for future progress.

% there is still a gap between the results of domain adaptation (\ie, without labeled data) and the supervised learning in the in-domain (\ie, with labeled data).
% \yi{previous sentence confusing}

% \zhiming{update} We also found that the improvement from Python to TypeScript is larger than the improvement from TypeScript to Python. The reason could be that 1) Python dataset is labeled with a more fine-grained way (\ie, 16 types) while TypeScript dataset is labeled more coarsely (\ie, 8 types), 2) the size of Python dataset is larger than the TypeScript dataset.The results tend to indicate that the in-domain dataset with higher quality (\eg, fine-grained labelling or more data) and diversity could achieve better performance on the out-domain dataset.

\begin{tcolorbox}[size=title,opacityfill=0.1,breakable]
\textbf{Answer to RQ1}:
\tool can effectively improve over the state-of-the-art domain adaptation methods and rule-based tool on NTL, \ie, the target language dataset is not labeled.
% in terms of exact match and weighted-F1 using either optionally-typed or strongly-typed languages as source domain.
\end{tcolorbox}

\begin{table}[t]
\centering
\caption{Results on the impact of different components.}\label{tab:ab}
\resizebox{.85\columnwidth}{!}{
\begin{tabular}{ccc|cc}
\toprule
\multirow{2}{*}{Methods} & \multicolumn{2}{c|}{Python $\rightarrow$ TypeScript} &
\multicolumn{2}{c}{TypeScript$\rightarrow$ Python}  \\
\cline{2-5}
                         & EM    & weighted-F1                      & EM    &
                         weighted-F1                     \\
\midrule
w/o SE                   & 0.740 & 0.703                             & 0.606 &
0.596                           \\
\midrule
w/o VTC               & 0.778 & 0.773                            & 0.618 & 0.610                         \\
\midrule
-Kernel                  & 0.730 & 0.712                           & 0.601 &
0.584                        \\
-Sequence               & 0.778 & 0.773                            & 0.618 & 0.610                          \\
\midrule
\tool     & 0.796 & 0.790                         & 0.633 & 0.618                           \\
\bottomrule
\end{tabular}}
\end{table}

\subsection{RQ2: Usefulness of Different Components}

\paragraph{Setup} In this section, we perform an ablation study to study the contribution of different components of our method in the results of $RQ1$. We build the following baselines to evaluate each component:
%\tool on Python $\rightarrow$ TypeScript and TypeScript $\rightarrow$ Python. We evaluate under the \textbf{NTL} setting and build the following baselines to evaluate each component:
\begin{itemize}[topsep=2pt,itemsep=2pt,partopsep=0ex,parsep=0ex,leftmargin=*]
    \item \textbf{\tool without syntax enhancement} (\textit{w/o SE}). We remove the component of syntax enhancement and let the neural network to learn the syntax mapping (\eg, different keywords) itself.
    \item \textbf{\tool without VTC-based Kernelized Attention} (\textit{w/o VTC}). We remove the VTC-based kernelized attention to evaluate its effect.
    % \item \textbf{\tool without Data-flow Analysis} (\textit{w/o VTC}). To evaluate the effect of kernelized attention, we removed the variable type closeness matrix from our system.
\item \textbf{Ensemble Inference}. To evaluate the usefulness of $\kappa\textit{-}$ bagging strategy, we use the unkernelized model (\textit{Sequence}) and the kernerlized model (\textit{Kernel}) to  perform the inference  separately.
\end{itemize}

% \begin{figure}
%   \centering
%   \begin{subfigure}{0.30\columnwidth}
%     \centering
%     \includegraphics[width=\columnwidth]{fig/baseurl_attn_bar.pdf}
%     \caption{Example 1}\label{fig:baseurl}
%   \end{subfigure}
%   \qquad
%   \begin{subfigure}{0.36\columnwidth}
%     \centering
%     \includegraphics[width=\columnwidth]{fig/showcolor_attn.pdf}
%     \caption{Example 2}\label{fig:showcolor}
%   \end{subfigure}
%   \caption{Attention visualization in which kernelized model is more accurate than unkernelized
%   model.}\label{fig:attnvis}
% \end{figure}

\paragraph{Usefulness of Syntax Enhancement}
\label{sec:rq3-results}
\looseness=-1
As shown in (Row \textit{w/o SE}) of \Cref{tab:ab}, the performance is significantly reduced compared with \tool. Specifically, after removing the meta-grammar representation, from Python to TypeScript, the results drop by 5.60\%@EM, 8.70\%@weighted-F1; while for TypeScript to Python, the results drop by 2.70\%@EM, 2.20\%@weighted-F1. The results indicate that by enhancing the input representation with meta-grammar representation, the overlapped features among language domains are significantly increased thus improving the transferability of the model.

\begin{figure}[t]
 %H为当前位置，!htb为忽略美学标准，htbp为浮动图形
\centering %图片居中
\includegraphics[width=0.4 \textwidth]{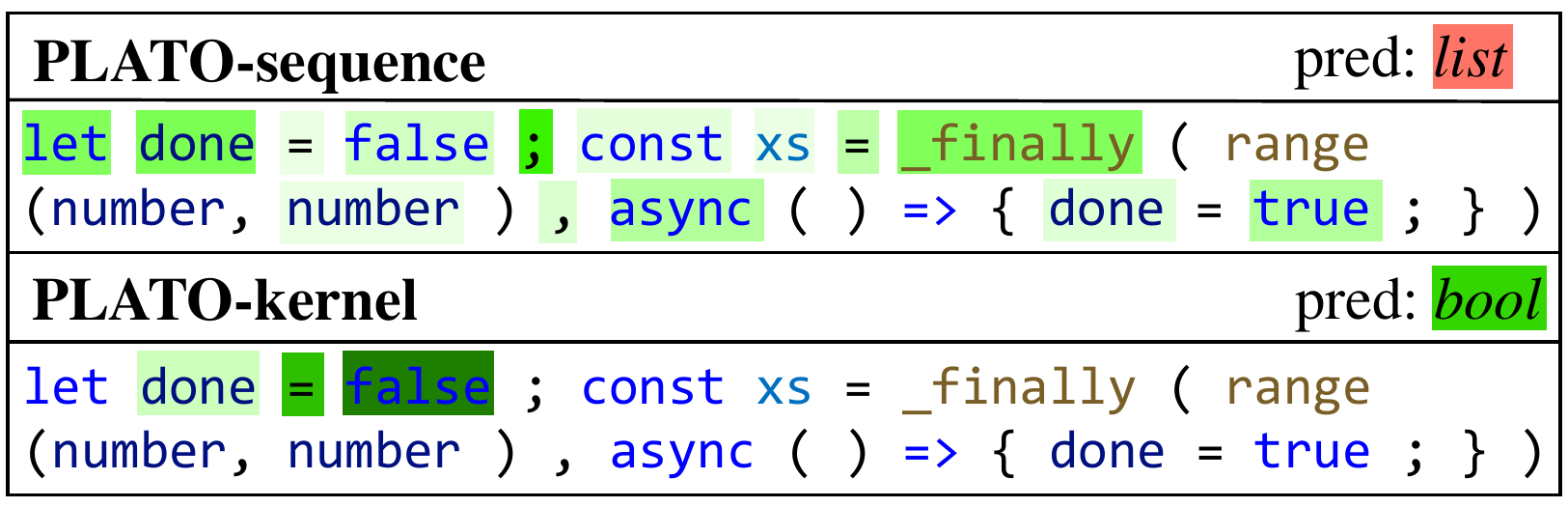} %插入图片，[]中设置图片大小，{}中是图片文件名
%最终文档中希望显示的图片标题
\caption{Illustrative example of kernelized model compared with original sequence model. The first and second row denote their attention vectors of the Boolean variable \texttt{done}.}
\label{fig:attn_rq1} %用于文内引用的标签
\end{figure}

\paragraph{Usefulness of Variable Type Closeness} \noindent
Consider the results in Row \textit{w/o VTC}, we found that the performance decreases in each task. Note that when removing VTC from the model, it degenerates into using the code sequence without the kernelized attention. Specifically, without VTC, from Python to TypeScript, the exact match and weighted-F1 are decreased by 1.80\% and 1.70\%, respectively. From TypeScript to Python, the performance is decreased by 1.50\% and 0.80\%, respectively. It demonstrates the usefulness of the VTC-based kernelized attention strategy.
\looseness=-1
We provide a case study to further demonstrate the usefulness of the VTC-based kernelized attention in \cref{fig:attn_rq1}. Specifically, we conduct max-pooling on the last multi-head self-attention layer to get the attention vector of the Boolean variable ``\texttt{done}'' for both the original sequence model and the kernelized model. The shade of the token denotes its attention weight. For the sequence model, it is shown that model spuriously leverages the irrelevant token ``\texttt{\_finally}'' for prediction while paying relatively low attention to the ground-truth evidence ``\texttt{false}'', and the variable is erroneously classified as list. And by incorporating the VTC-based kernelized attention, the model robustly infers the variable as Boolean based only on the ground-truth evidence. The visualization shows that VTC-based kernelized attention forces the model to base its inference on relevant, domain-invariant features thus makes it more robust and transferable among language domains.

\paragraph{Impact of Ensemble-based Inference} Rows ``-$\mathrm{Seuquence}$'' and ``-$\mathrm{Kernel}$'' show that \tool substantially outperforms the two
sub-models. Note that the results of \tool w/o VTC and \tool-Sequence are the same because the sequence model is the version of \tool without the VTC-based kernelized attention. The $\kappa\textit{-}$ bagging ensemble strategy can make the best of the kernelized model and compensate its weakness when dealing with language-specific corner cases.

\begin{figure*}[t]
\centering
  % Maximum length
 \subcaptionbox{Py $\rightarrow$ TS intra
 EM\label{js_intra_acc}}{
 \includegraphics[width=.25\textwidth]{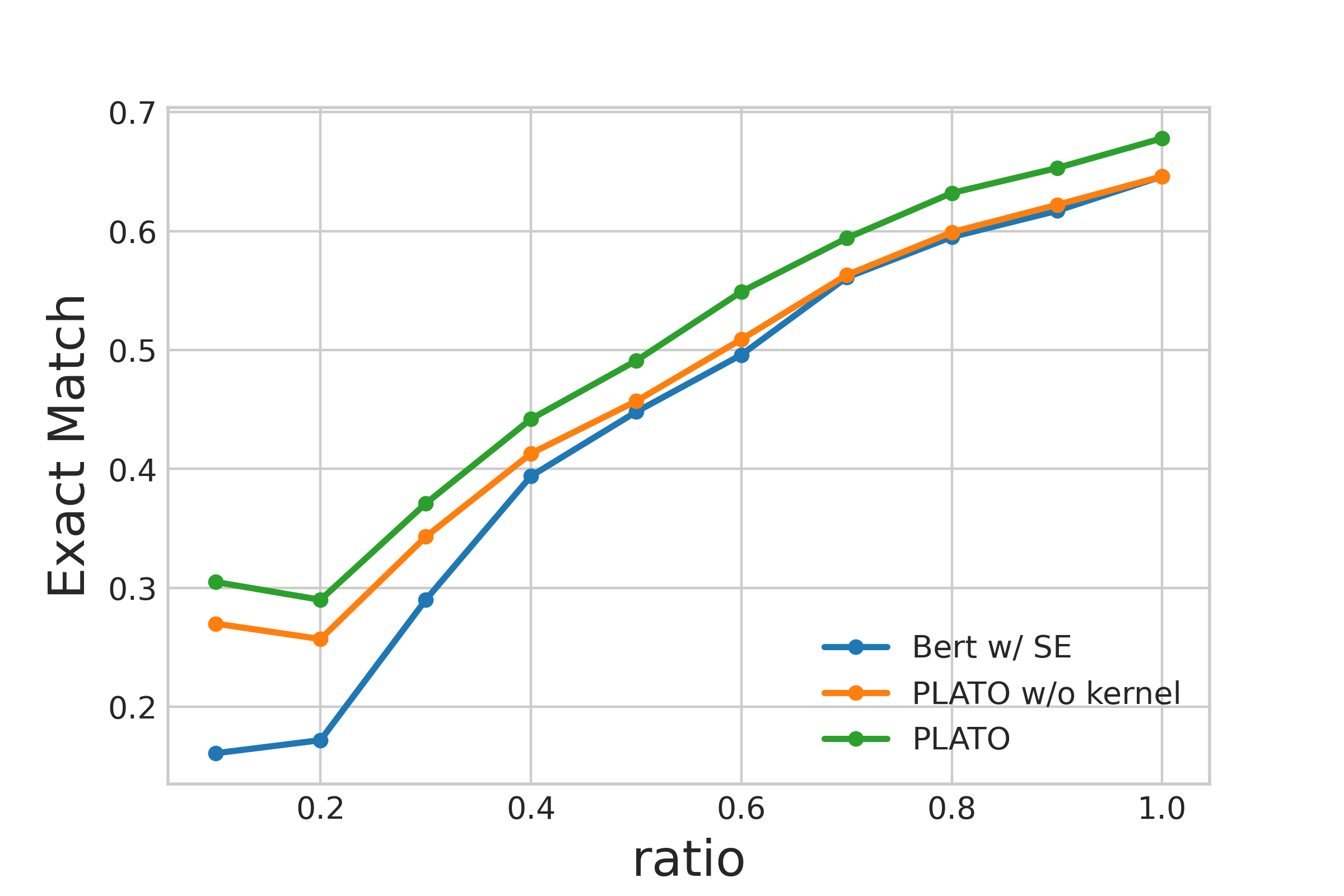}}%
 \subcaptionbox{Py$\rightarrow$ TS intra
 weighted-F1\label{js_intra_f1}}{
 \includegraphics[width=.25\textwidth]{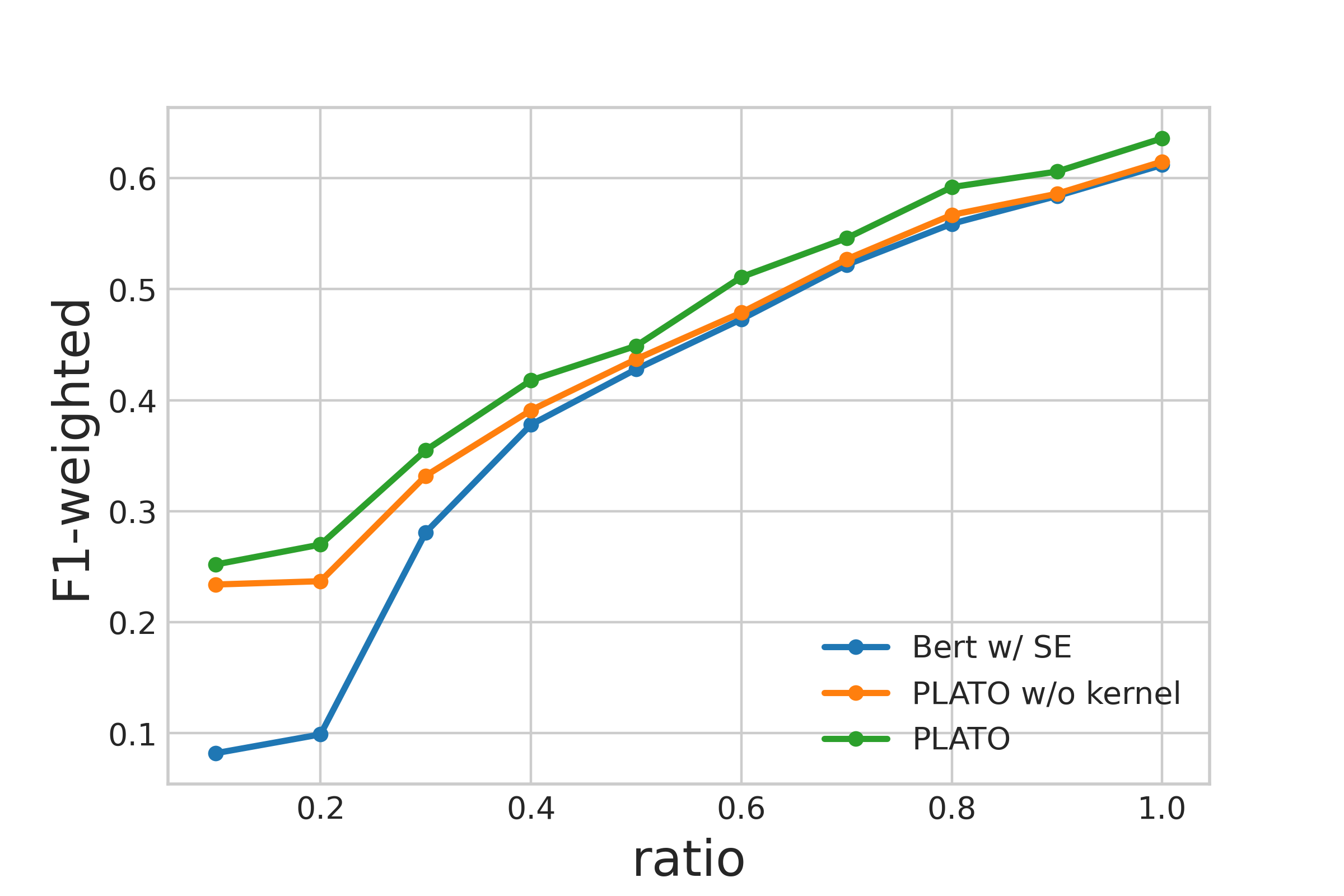}}%
  \subcaptionbox{Py $\rightarrow$ TS inter
  EM\label{js_inter_acc}}{
  \includegraphics[width=.25\textwidth]{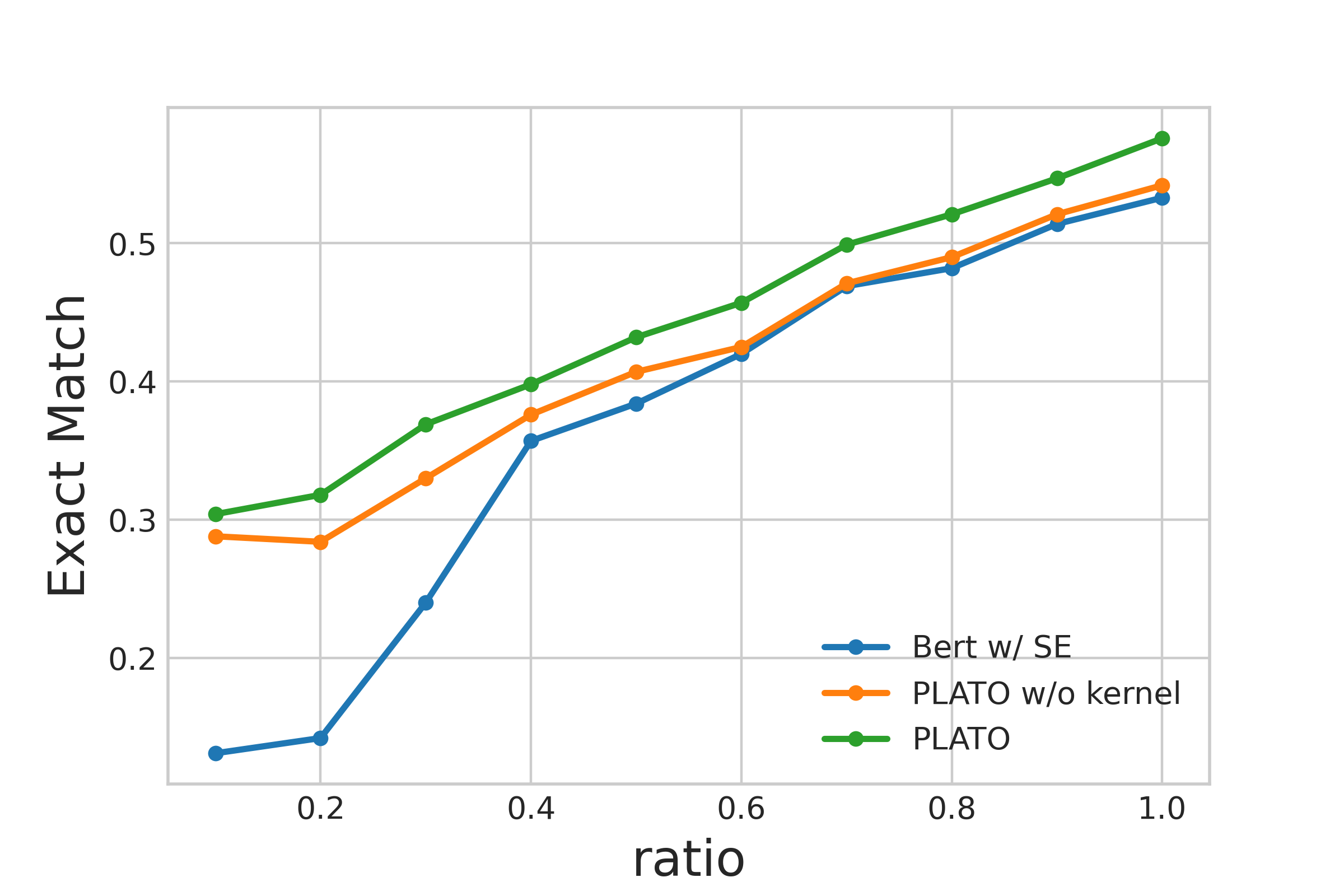}}%
  \subcaptionbox{Py $\rightarrow$ TS inter
  weighted-F1\label{js_inter_f1}}{
  \includegraphics[width=.25\textwidth]{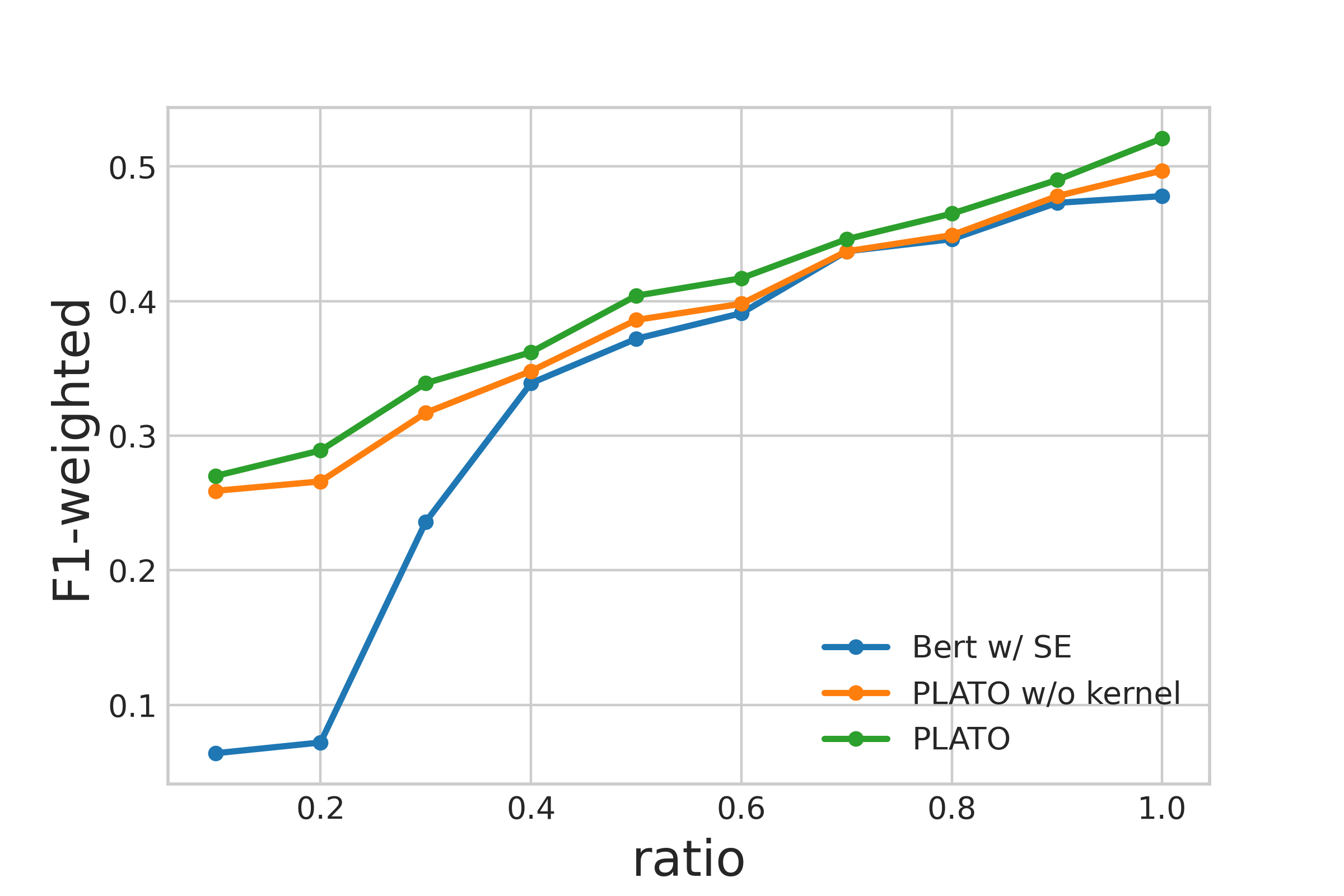}}%

 \subcaptionbox{TS $\rightarrow$ Py intra EM\label{py_intra_acc}}{
   \includegraphics[width=.25\textwidth]{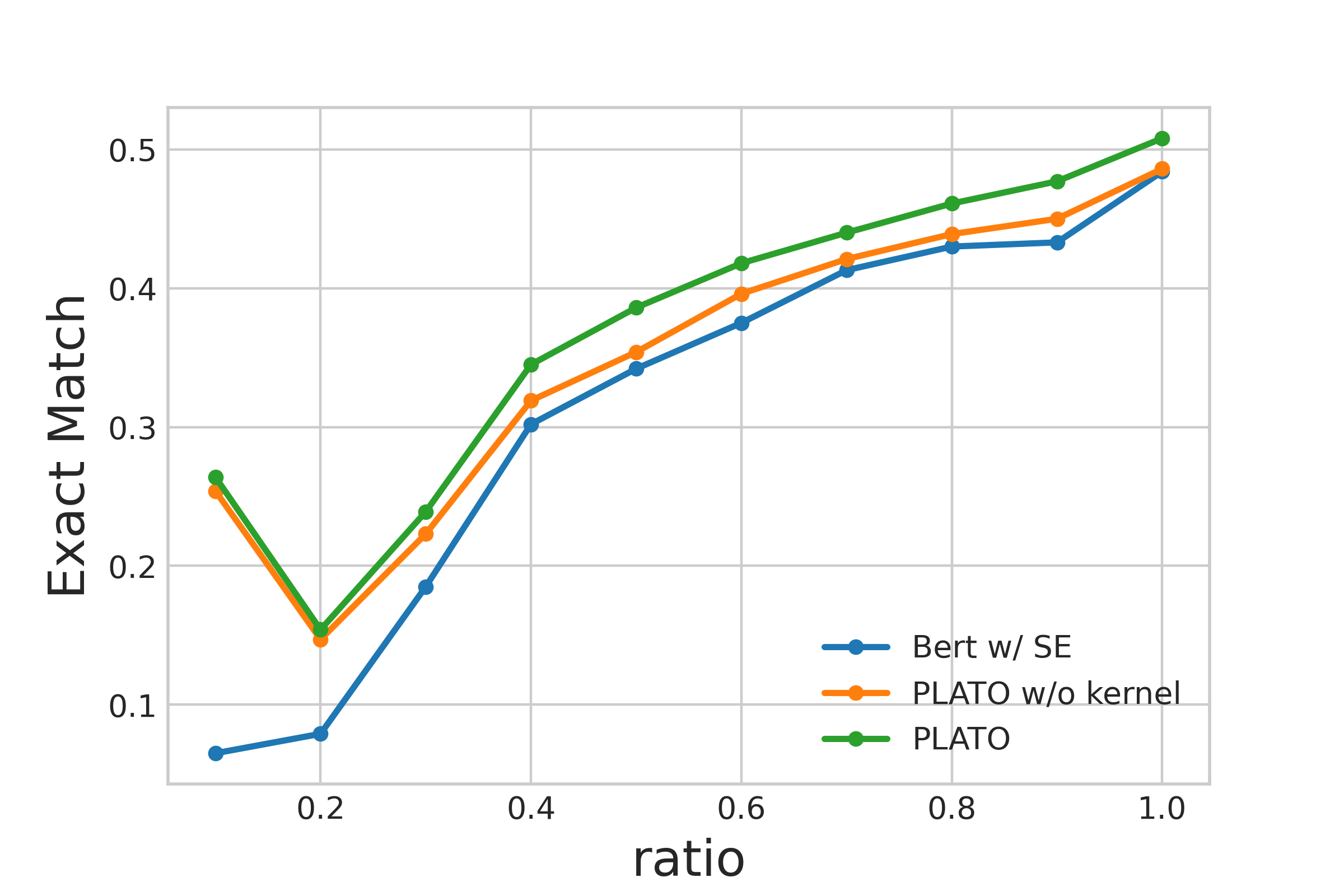}}%
  \subcaptionbox{TS $\rightarrow$ Py intra weighted-F1\label{py_intra_f1}}{
    \includegraphics[width=.25\textwidth]{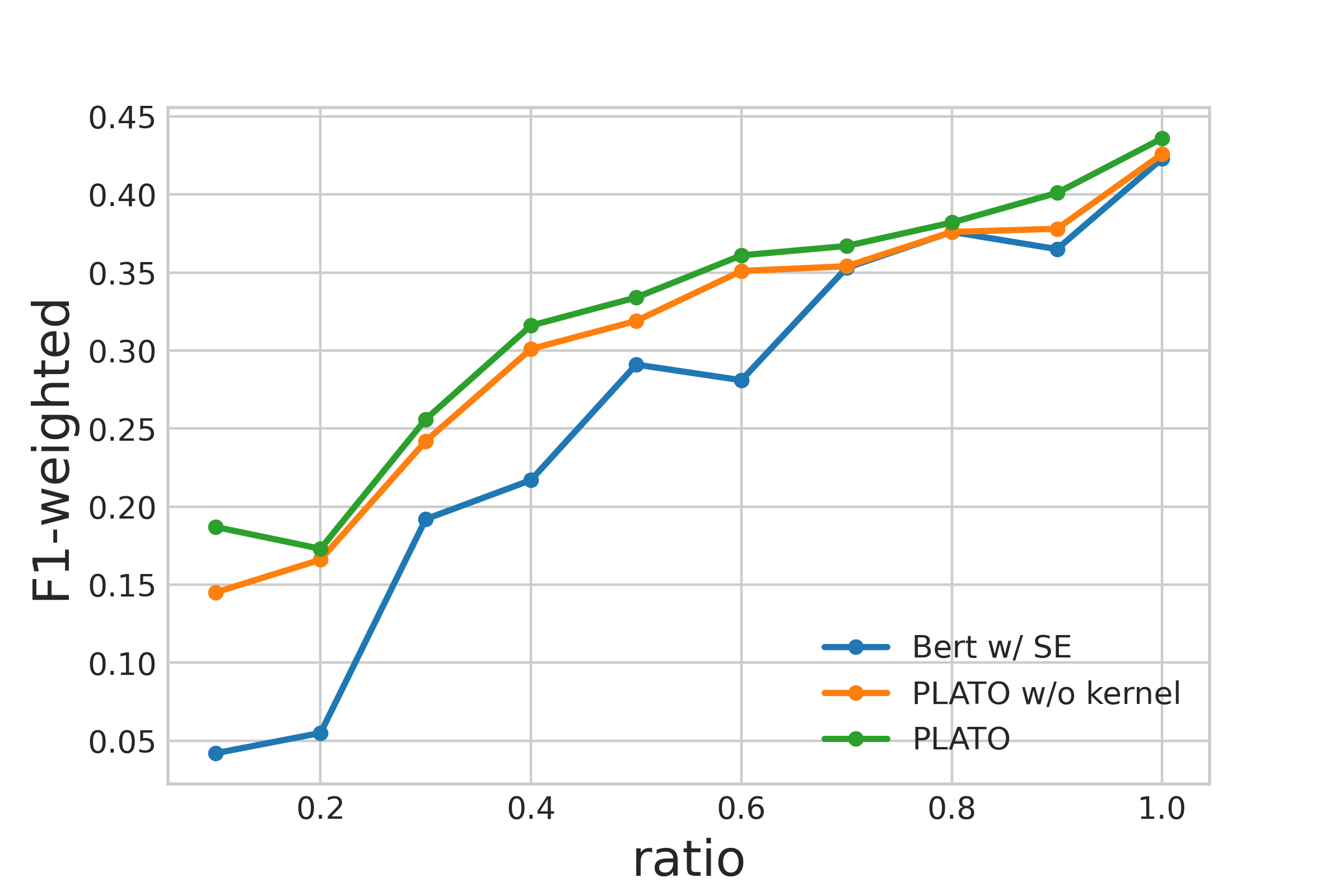}}%
  \subcaptionbox{TS $\rightarrow$ Py inter EM\label{py_inter_acc}}{
    \includegraphics[width=.25\textwidth]{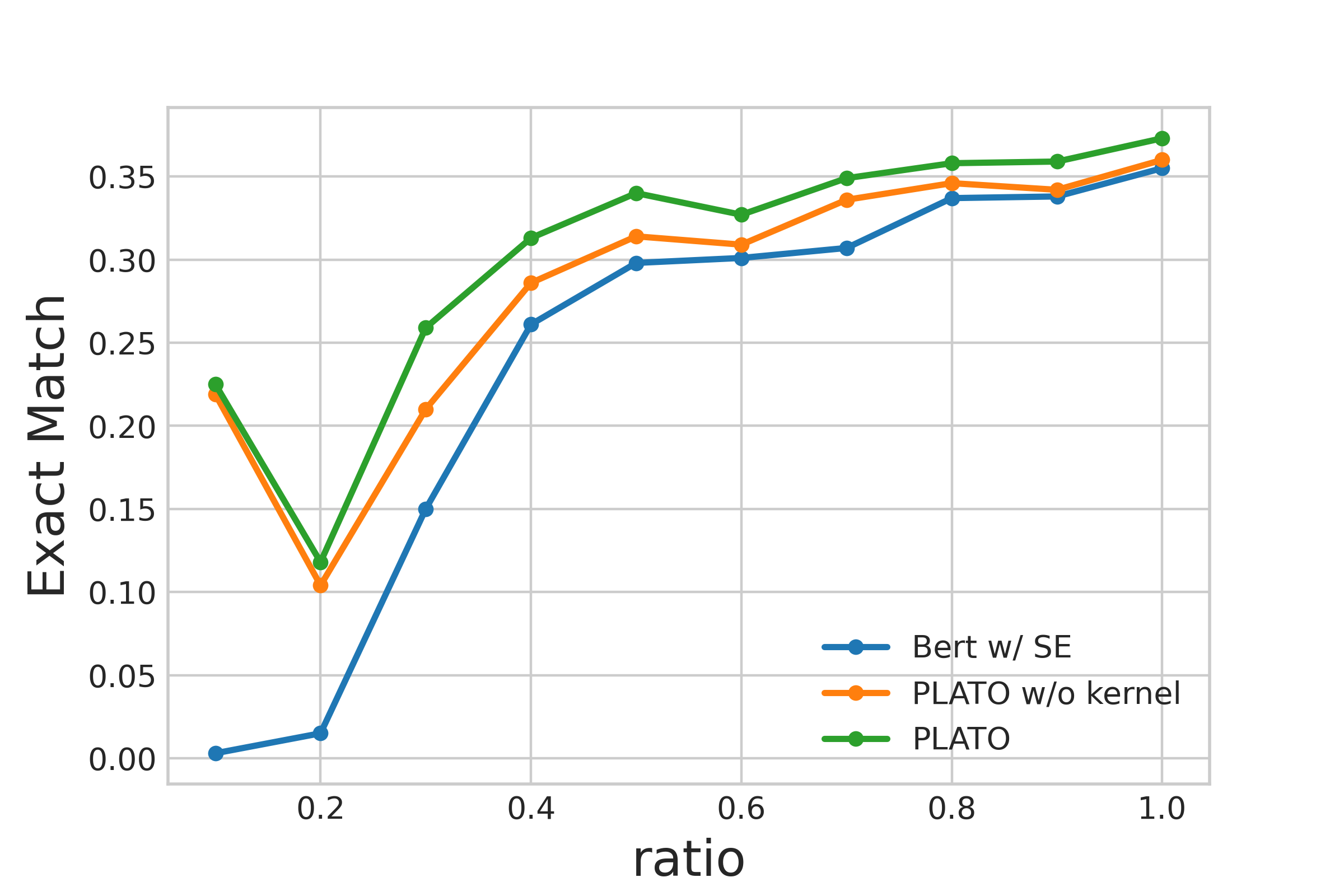}}%
  \subcaptionbox{TS $\rightarrow$ Py inter weighted-F1\label{py_inter_f1}}{
    \includegraphics[width=.25\textwidth]{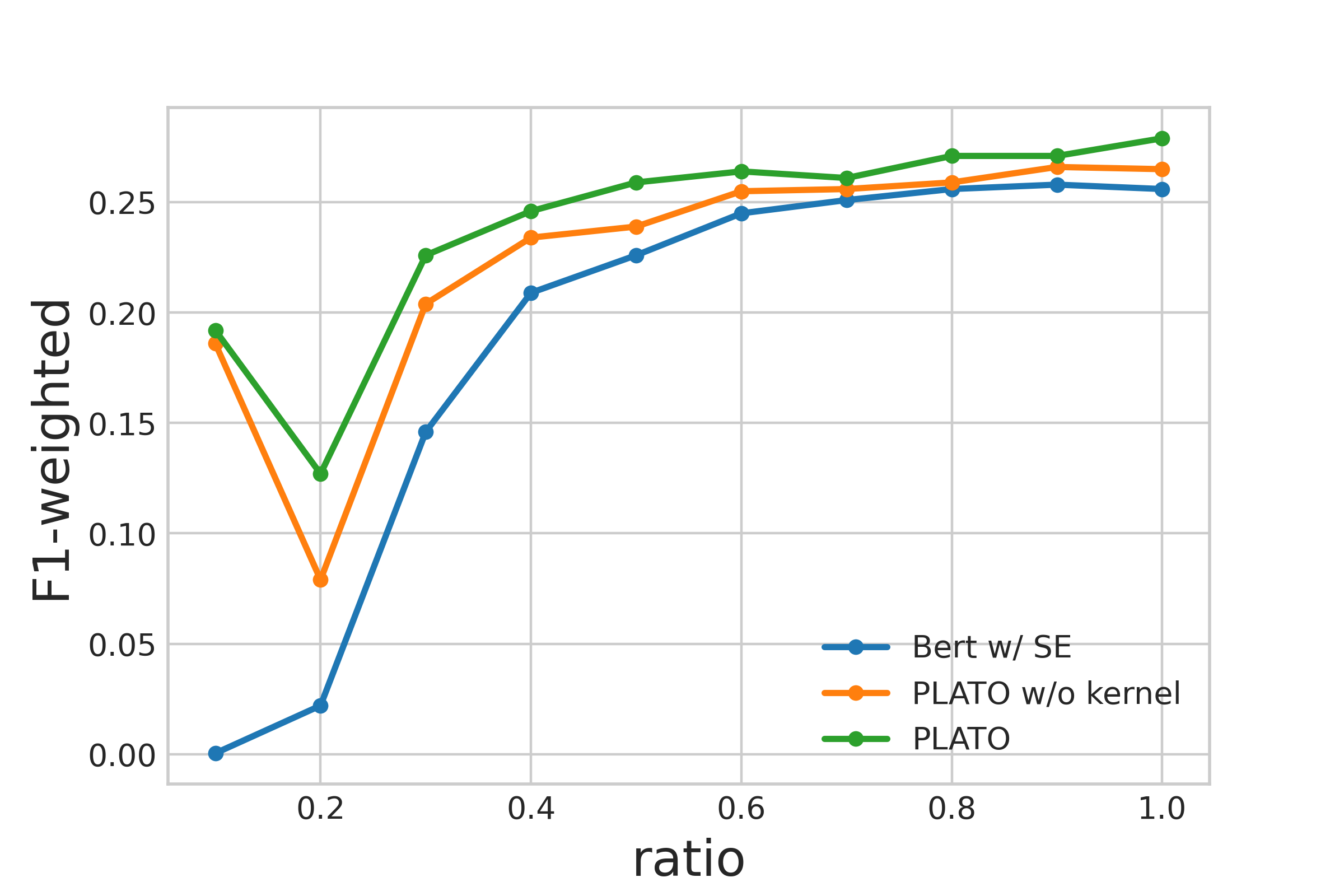}}%
  \caption{The evaluation results when partial labeled data is available.}\label{fig:ava}
\end{figure*}

% \begin{table}[t]
% \caption{Comparison of \tool's performance with baseline methods on TypeScript dataset}
% \begin{tabular}{ccc|cc}
% \hline
% \multirow{2}{*}{Method} & \multicolumn{2}{c|}{intra-project} & \multicolumn{2}{c}{inter-project} \\ \cline{2-5}
%           & EM     & weighted-F1 & EM     & weighted-F1 \\ \hline
% LambdaNet & 0.646  & 0.623       & 0.535  & 0.481       \\
% Transformer   & 0.695  & 0.654       & 0.532  & 0.484       \\
% TypeBert  & 0.723  & 0.698       & 0.551  & 0.504       \\
% \tool                   & \textbf{0.760}   & \textbf{0.729}  & \textbf{0.567}  & \textbf{0.529}  \\ \hline
% $\Delta$     & +3.70\% & +3.10\%      & +1.60\% & +2.50\%      \\ \hline
% \end{tabular}

% \label{2js_full}
% \end{table}

\begin{tcolorbox}[size=title,opacityfill=0.1,breakable]
\textbf{Answer to RQ2}:
Each component in \tool is useful for the cross-lingual transfer learning of statistical type inference task. In conclusion, syntax enhancement improves the performance significantly by introducing feature overlap among language domains. The VTC-based kernelized attention mechanism improves performance consistently by forcing model to pay attention to relevant, domain-invariant features.
\end{tcolorbox}
\looseness=-1
\subsection{RQ3: Using Partial Labeled Target Language Data}

% % full table2 (python)
% \begin{table}[t]
% \caption{Comparison of \tool's performance with baseline methods on Python dataset}
% \begin{tabular}{ccc|cc}
% \hline
% \multirow{2}{*}{Method} & \multicolumn{2}{c|}{intra-project} & \multicolumn{2}{c}{inter-project} \\ \cline{2-5}
%          & EM     & weighted-F1 & EM     & weighted-F1 \\ \hline
% Typilus  & 0.516  & 0.484       & 0.441  & 0.412       \\
% Transformer     & 0.463  & 0.372       & 0.431  & 0.425       \\
% TypeBert & 0.522  & 0.490        & 0.435  & 0.428       \\
% \tool                   & \textbf{0.559}   & \textbf{0.514}  & \textbf{0.482}  & \textbf{0.447}  \\ \hline
% $\Delta$    & +3.70\% & +2.40\%      & +4.10\% & +1.90\%      \\ \hline
% \end{tabular}

% \label{2py_full}
% \end{table}

% \usepackage{multirow}
% \usepackage{booktabs}

\paragraph{Setting} In the real-world settings, during the early stage of an optionally-typed programming language, the type hint annotations of the language provided by developers are scarce, especially for primitive types (\eg~for the TypeScript dataset, without the data augmentation of the CheckJS tool, 70.7\% of samples do not contain the overlapped meta-types, while only 22.9\% after augmentation). Thus, it would be extremely valuable if we were able to quickly build a functional type inference tool by leveraging existing cross-lingual labeled dataset to augment the training data of the model during the early stage of a language. To simulate the early stage, \ie~without augmentation of the existing tool, we sort the sample order of the dataset (post-augmented) according to the number of overlapped meta-types within each sample from least to most. And we select 10\%, 20\%, \ldots, 100\% chunks of samples from the target dataset together with 10\% of the known source language dataset to train the model. Note that we only select a few source language data (\ie, 10\%) because we try to reduce the effect of the size of the source data on the final results. We use 5,000 samples for both the source and target dataset. The following baselines are selected to demonstrate the usefulness of \tool:
\begin{itemize}[topsep=2pt,itemsep=2pt,partopsep=0ex,parsep=0ex,leftmargin=*]
    % \item \textit{State-of-the-art tools}. We compare \tool with previous state-of-the-art learning based inferenece tool. For TypeScript, we select TypeBert\cite{jesse2021learning}, LambdaNet\cite{wei2020lambdanet} and the vanilla Transformer model. For Python, we select TypeBert, Typilus and Transformer. For these supervised baselines\fei{what are these? need to be clear}, we use 100\% target dataset with labels. \fei{Check whether it belongs tow RQ4. Check the consistency of RQ3 and RQ4. }

    % we selected a static type inference tool CheckJs\footnote{https://github.com/Microsoft/TypeScript/wiki/Type-Checking-TypeScript-Files} and DeepTyper~\cite{hellendoorn2018deep} which is the state-of-the-art deep learning based type inference tool for TypeScript. For Python, we tried our best to configure the baselines (\eg, \cite{allamanis2020typilus}) but most of them are not available in our evaluation. For the supervised baselines, we use 100\% target data.
    %we did not compare with the state-of-the-art baseline\cite{allamanis2020typilus} since their work focuses on open-world types using metric learning while ours focused on closed-world types and it cannot be put under fair comparison.
    \begin{table}
\centering
\caption{Analysis of \tool's performance on the dataset with different sizes.}
\begin{tabular}{ccc|cc} 
\toprule
\multirow{2}{*}{Method} & \multicolumn{2}{c|}{$Ts_{10k}$} & \multicolumn{2}{c}{$Py_{10k}$}  \\
                        & EM    & weighted-F1             & EM    & weighted-F1              \\ 
\cmidrule(lr){1-5}
BERT w/ SE               & 0.700 & 0.663                   & 0.517 & 0.464                    \\
PLATO w/o kernel        & 0.702 & 0.677                   & 0.528 & 0.474                    \\
PLATO                   & 0.730 & 0.696                   & 0.553 & 0.490                     \\
\bottomrule
\end{tabular}
\label{10k_exp}
\end{table}

    \item \textit{Bert with SE}. Since our method is based on Bert, to demonstrate the effectiveness of cross-lingual data augmentation, we fine-tune the pretrained XPLM model on the partial labeled target language data with syntax enhancement with the fully-supervised learning paradigm.
    \item \textit{\tool without kernel}. To show the effect of the kernelized attention on PTL, we evaluate the unkernelized sub-model of \tool, \ie, removing the kernelized model from \tool.
\end{itemize}

We follow the settings in previous works~\cite{allamanis2020typilus,hellendoorn2018deep} and evaluate the results under two settings:
(1) \textit{intra project}: the training and test dataset come from same project sets;
(2) \textit{inter project}: the training and test dataset come from different project sets.

\begin{figure*}[t]
 %H为当前位置，!htb为忽略美学标准，htbp为浮动图形
\centering %图片居中
\includegraphics[width=0.85\textwidth]{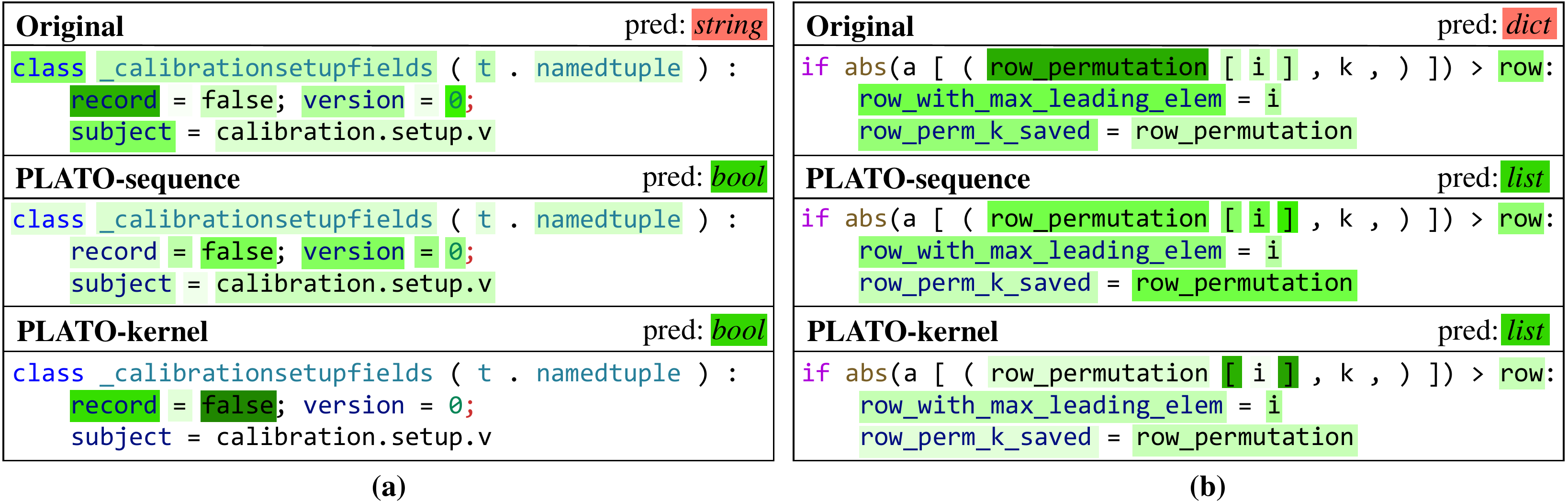} %插入图片，[]中设置图片大小，{}中是图片文件名
%最终文档中希望显示的图片标题
\caption{Illustrative examples of the attention vector of the BERT w/ SE, \tool w/o kernel and \tool-kernel models. (a) and (b) show the attention vectors of the Boolean variable \texttt{record} and the list variable \texttt{row\_permutation}. The attention vector is much more unbiased and transferable with the incorporation of cross-lingual data and kernelized attention.}
\label{fig:attnvis} %用于文内引用的标签
\end{figure*}

\paragraph{Results} \Cref{fig:ava} shows the results.  First, we can see that \tool w/o kernel steadily outperforms the baseline Bert model under both settings. Particularly, the improvement is more significant when the size of the target labeled data is small ($\mathrm{ratio<0.5}$). For example, under the intra-project setting for $\mathrm{Python \rightarrow TypeScript (intra)}$, when $\mathrm{ratio=0.1}$, it improves the baseline model by +10.90\%@EM and +15.20\%@weighted-F1. Besides, cross-lingual data augmentation is still useful when using full target training set ($\mathrm{ratio=1.0}$), \eg~for $\mathrm{TypeScript \rightarrow Python (inter)}$, it improves the baseline Bert by +0.50\%@EM and +0.90\%@weighted-F1. The results demonstrate the effectiveness of out-domain cross-lingual data augmentation. Then, we consider the results of \tool,  as
shown in \cref{fig:ava}, \tool significantly improves over Bert and \tool w/o kernel
under all ratios of labeled target language data for both the intra- and inter-project settings. For example, for $\mathrm{Python \rightarrow TypeScript (intra)}$, when $\mathrm{ratio=0.1}$, \tool further increases the \textit{\tool without kernel} baseline by +3.50\%@EM
and +1.80\%@weighted-F1; when $\mathrm{ratio=1.0}$, \tool manages to increase it by +3.20\%@EM
and +2.10\%@weighted-F1. The result is consistent for $\mathrm{TypeScript \rightarrow Python}$.
\begin{table}
\centering
\caption{Analysis of models' attention score with cross-lingual data augmentation and kernelized attention.}
\begin{tabular}{cccc} 
\toprule
\multirow{2}{*}{Type} & \multicolumn{3}{c}{attention score $(\times10^{-3})$}  \\
                      & BERT w/ SE & PLATO w/o kernel & PLATO-ker                \\ 
\cmidrule(lr){1-4}
Bool                  & 4.95 & 6.06      & 566.00                      \\
List                  & 0.82 & 3.65      & 10.01                    \\
\bottomrule
\end{tabular}
\label{attn_full}
\end{table}
The results indicate that using our kernerlized attention can further boost the performance. We also conduct experiments on the dataset with 10,000 samples (denoted as $Ts_{10k}$ and $Py_{10k}$) under the intra-project setting, the results are shown in \cref{10k_exp}. It is obvious that cross-lingual data augmentation and the kernelized attention are also effective even when the target language data is very sufficient. \Eg~for $Py_{10k}$, \textit{\tool without kernel} improves the baseline BERT model by +1.10\%@EM
and +1.00\%@weighted-F1; and \tool further improves the \textit{\tool without kernel} by +2.50\%@EM
and +1.60\%@weighted-F1. In order to understand the source of improvement, we conduct case studies and quantitative analysis of the three models ($Py_{10k}$) as shown in \cref{fig:attnvis} and \cref{attn_full}. We conduct max-pooling on the last multi-head self-attention layer to get the attention score. \cref{fig:attnvis} shows the attention visualization of the Boolean variable \texttt{record} and the list variable \texttt{row\_permutation}. \Eg~for \texttt{record}, the original model spuriously pays high attention to the irrelevant identifier \texttt{subject} while low attention to the ground-truth evidence \texttt{false} and thus erroneously infers the variable as \texttt{string}. With the cross-lingual data augmentation, the \textit{\tool without kernel} model pays less attention to the irrelevant tokens and more attention to \texttt{false} and thus infers correctly. Finally, with kernelized attention, the \tool-kernel model completely ignores irrelevant tokens and correctly infers with only the ground-truth evidence.
% full table1
\begin{table}[t]
\caption{Comparison of \tool's performance with baseline methods on the TypeScript dataset}
\centering
\begin{tabular}{ccc} 
\toprule
Method               & EM (Top-1) & EM (Top-5)  \\ 
\cmidrule(r){1-3}
LambdaNet (lib only) & 0.770      & -           \\
DeepTyper (lib only) & 0.674      & -           \\
DeepTyper (all)      & 0.569      & 0.811       \\
PLATO-seq (all)      & 0.679      & 0.821       \\
PLATO (all)          & 0.716      & 0.838       \\
\bottomrule
\end{tabular}
\label{2js_full}
\end{table}
Furthermore, as shown in the quantitative analysis shown in \cref{attn_full}, we compute the mean attention score of the ground-truth evidence. for Boolean variables, we focus on the assignment statements within which the literal \texttt{true} and \texttt{false} are ground truth. And for list variables, we focus on the list indexing, we compute the attention score of the corresponding indexing bracket \texttt{[} of the variables. The results in \cref{attn_full} indicate that the attention score of ground-truth evidence is increased with the cross-lingual source data augmentation for both the Boolean and list variables. And with the introduction of kernelized attention mechanism, the attention score is further boosted. 

\begin{table}[t]
\caption{Comparison of \tool's performance with baseline methods on the Python dataset}
\centering
\begin{tabular}{ccc} 
\toprule
Method               & EM    & EM (parametric)  \\ 
\cmidrule(lr){1-3}
Typilus (graph2class) w/ pytype & 0.461 & 0.488            \\
Typilus w/o pytype              & 0.502 & 0.575            \\
PLATO-seq w/o pytype            & 0.509 & 0.571            \\
PLATO w/o pytype                & 0.546 & 0.607            \\
\bottomrule
\end{tabular}
\label{2py_full}
\end{table}
\begin{tcolorbox}[size=title,opacityfill=0.1,breakable]
\textbf{Answer to RQ3}: As more labeled target language data is available, the performance of \tool is steadily increased and it outperforms the baseline models consistently under all ratios of target language data by leveraging more unbiased and transferable features.
\end{tcolorbox}
\looseness=-1
\subsection{RQ4: Evaluation on Supervised Learning}
\paragraph{Setting} In this evaluation, we applied \tool in the fully supervised learning scenario to evaluate whether our model manages to improve over previous baseline methods. 
For the supervised baselines, we select the following state-of-the-art baselines: for TypeScript, we compare \tool with LambdaNet~\cite{wei2020lambdanet} and the DeepTyper model~\cite{hellendoorn2018deep}. For Python, we compare with the Typilus model~\cite{allamanis2020typilus}. For the TypeScript baselines, we use the dataset provided by DeepTyper following previous RQs. we follow the evaluation setting of DeepTyper and randomly select 10\% projects for testing, 10\% for validation and 80\% for training. For the Python baselines, we follow the same intra-project data split setting of Typilus and randomly split the data into train-validation-test set in 70-10-20 proportions. We report the results of the baselines from the original paper and GitHub repositories. We evaluate all models with the same measurements used in the original paper of the baseline models. Table~\ref{2js_full}, \ref{2py_full} show the results on TypeScript and Python, respectively. 

For TypeScript, we report the original results of LambdaNet (lib only) and DeepTyper (lib only) for reference, which are evaluated on the LambdaNet dataset that contains library type annotations only. And we report the original results of DeepTyper (all) and \tool (all) which are evaluated on the DeepTyper dataset with a much larger prediction space that contains both the user-defined and library type annotations. It is obvious that \tool (all) achieves better results than DeepTyper (all) and \tool-seq (all) (\tool w/o kernelized attention) due to the kernelized attention mechanism. Specifically, \tool (all) improves over \tool-seq (all) by 3.70\%@EM(Top-1) and 1.70\%@EM(Top-5). And for Python, our \tool-seq w/o pytype achieves similar performance with Typilus w/o pytype. By incorporating the kernelized attention mechanism, the full \tool model manages to improve the performance of Typilus baseline by +4.40\% (EM) and +3.20\% (EM up to parametric type).

\begin{tcolorbox}[size=title,opacityfill=0.1,breakable]
\textbf{Answer to RQ4}: Under the same evaluation settings, \tool manages to outperform the compared baseline methods on both the TypeScript and Python datasets.
% \tool is useful in improving the performance of the existing supervised-learning based methods via the domain adaptation on a small amount of out-domain cross-lingual data.
\end{tcolorbox}

\subsection{Threats to Validity}
The implementation of the baselines is a threat to the validity of the results.
Since these techniques were not originally built for program analysis tasks, we gave our best
efforts in adapting them for our tasks, and fixed all bugs we could identify.
The selection of the datasets may not be representative and our results may not generalize.
To mitigate this, we selected the two well-known benchmarks which were previously used in type inference tasks.
% The selection of hyper-parameters could be another threat and we mitigate it by
Finally, the label calibration (see Section~\ref{sec:setup}) could be another threat to the accuracy of the type
prediction.
This can be mitigated by outputting specific type names within a meta-type in a ranked list to developers. 

\section{Related work}

\subsection{Unsupervised Domain Adaptation}
As an important case of transfer learning, unsupervised domain adaptation (UDA) has drawn 
significant attention from the deep learning communities. 
The UDA research can mainly be categorized into two streams~\cite{ramponi2020neural}, namely model-centric and data-centric. The goal of model-centric methods are to minimize the distance among domains via feature alignment.
Tzeng et al.~\cite{tzeng2014deep} first proposed using the maximum mean discrepancy (MMD) to 
minimize the distance between images from two distributions on image classification tasks. 
Recently, NLP community also started to investigate the possibility of applying the above 
techniques to language tasks, \eg, sentiment classification~\cite{li2018s, shen2018wasserstein}, POS 
tagging~\cite{yasunaga2017robust}, etc. 
Pan et al.~\cite{pan2010cross} proposed spectral feature alignment for sentiment classification; the syntax enhancement approach we used in this work lies in this category. The goal of data-centric methods are to bridge the domain gap by manipulating data from source 
and target domains. Han et al.~\cite{han2019unsupervised} proposed the \textit{adaptive pre-training}, 
which adapts contextualized word embeddings from target domain by masked language modeling. 
Gururangan et al.~\cite{gururangan2020don} further introduced \emph{task-specific pre-training} (TAPT) that studies the effect of second-stage pre-training on the transferability across domains. In the software engineering community, transfer learning techniques 
started to gain attention recently. 
Nam et al.~\cite{nam2013transfer} proposed using transfer learning to improve the performance for cross-project defect prediction. SAR~\cite{bui2019sar} leverages generative adversarial network for API mappings. Although UDA has been broadly explored in the CV and NLP fields, it has not been paid enough
attention in the programming language and software engineering community. 
Yet, considering the fact that we have abundant labeled dataset for high-resource programming 
languages, there is great potential for knowledge transfer to the relatively low-resource programming languages via UDA.

\subsection{Statistical Type Inference}
Type inference for optionally-typed language is widely studied in light of the widespread usage of 
languages such as Python and JavaScript. 
The ability to infer types automatically makes programming tasks easier, leaving the 
programmer free to omit annotations while still permitting type checking.
Statistical type inference is gaining attention due to its superior performance over traditional
rule-based methods.
JSNice~\cite{raychev2015predicting} proposed the first probabilistic type inference system based
on conditional random fields (CRFs). 
\textsc{DeepTyper}~\cite{hellendoorn2018deep} introduced the first deep learning based
JavaScript type inference model based on recurrent neural networks. And TypeBert~\cite{jesse2021learning} achieves the state-of-the-art performance thanks to unsupervised pre-training.
Following this line, several deep learning based type inference tools for Python are proposed~\cite{allamanis2020typilus, pradel2020typewriter}. \tool advances over these works by allowing the deep learning models to still work even without adequate labeled data.

\subsection{Program Representation Learning}
Leveraging deep learning models for solving software engineering problems is increasingly 
gaining popularity. 
Most of these works focus on monolingual tasks. 
Zhang et al.~\cite{zhang2020retrieval} used a recurrent neural network for code summarization in Python; 
code2vec~\cite{alon2019code2vec} used an attention model for method name prediction in Java; Graph neural networks~\cite{zhou2019devign,zhou2021spi} have been used for the vulnerability detection and security patches tasks of C.
% 给几个例子的work, devign啥的
Recently, researches started to investigate the power of multi-lingual language models for 
program analysis tasks.
Transcoder~\cite{lachaux2020unsupervised} introduced a neural transcompiler that is able to 
translate functions between C++, Java, and Python using unsupervised machine translation.
\section{Conclusion}
In this work, we set out to conduct the first trial of cross-lingual transfer learning of statistical type inference. Our experimental results are positive: by incorporating graph kernel-based kernelized attention, incorporating syntax enhancement using meta-grammar. Our framework not only improves previous domain adaptation baselines significantly when no labeled target language data is available, but also manages to consistently improve the supervised baseline when labeled target language data is available. Our findings indicate great potential of leveraging data across different programming languages for other neural model architectures and other different deep learning-based software engineering tasks. In the future, we plan to extend our method to more code learning based applications such as code search~\cite{gu2018deep} and code summarization~\cite{liu2021retrievalaugmented}, and improve the quality of the trained models with existing techniques~\cite{xie2019deephunter,du2019deepstellar,xie2021rnnrepair}. 

\begin{acks}
This research is partially supported by the National Research Foundation, Singapore under its the
AI Singapore Programme (AISG2-RP-2020-019), the National Research Foundation, Prime Ministers
Office, Singapore under its National Cybersecurity R\&D Program (Award No. NRF2018NCR-NCR005-0001),
NRF Investigatorship NRF-NRFI06-2020-0001, the National Research Foundation through its National
Satellite of Excellence in Trustworthy Software Systems (NSOE-TSS) project under the National
Cybersecurity R\&D (NCR) Grant award no. NRF2018NCR-NSOE003-0001, the Ministry of Education,
Singapore under its Academic Research Fund Tier 1 (21-SIS-SMU-033), Tier 2 (MOE2019-T2-1-040) and
Tier 3 (MOET32020-0004). Any opinions, findings and conclusions or recommendations expressed in
this material are those of the author(s) and do not reflect the views of the Ministry of Education,
Singapore.
\end{acks}

\balance
\bibliographystyle{plain}
\bibliography{main}

\end{document}